\documentclass[acmsmall]{acmart}
\AtBeginDocument{%
  \providecommand\BibTeX{{%
    \normalfont B\kern-0.5em{\scshape i\kern-0.25em b}\kern-0.8em\TeX}}}

\setcopyright{acmcopyright}
\copyrightyear{2023}
\acmYear{2023}
\acmDOI{XXXXXXX.XXXXXXX}

\acmJournal{CSUR}
\acmArticle{1}
\acmMonth{1}

\usepackage{makecell}
\usepackage{subfigure}
\usepackage{bm}
\usepackage{multirow}
\usepackage{multicol}
\usepackage{tabularx}
\usepackage{booktabs}
\usepackage{bbding}
\usepackage{cleveref}
\usepackage{hyperref}
\usepackage{colortbl}
\makeatletter
\definecolor{mygray}{gray}{.9}
\newcommand\figcaption{\def\@captype{figure}\caption}
\newcommand\tabcaption{\def\@captype{table}\caption}
 
\makeatother

\begin{document}

\title{A Survey of Neural Trees}


\author{Haoling Li}
\affiliation{%
  \institution{Zhejiang University}
  \country{China}
  }
  
\author{Jie Song}
\affiliation{%
  \institution{Zhejiang University}
  \country{China}}
  
\author{Mengqi Xue}
\affiliation{%
  \institution{Zhejiang University}
  \country{China}
  }
  
\author{Haofei Zhang}
\affiliation{%
  \institution{Zhejiang University}
  \country{China}
  }
  
\author{Jingwen Ye}
\affiliation{%
  \institution{National University of Singapore}
  \country{Singapore}
  }
  
\author{Lechao Cheng}
\affiliation{%
  \institution{Zhejiang Lab}
  \country{China}
  }
  
\author{Mingli Song}
\affiliation{%
  \institution{Zhejiang University}
  \country{China}
  }

\renewcommand{\shortauthors}{Li, et al.}

\begin{abstract}
Neural networks (NNs) and decision trees (DTs) are both popular models of machine learning, yet coming with mutually exclusive advantages and limitations. To bring the best of the two worlds, a variety of approaches are proposed to integrate NNs and DTs explicitly or implicitly. In this survey, these approaches are organized in a school which we term as neural trees (NTs). This survey aims to present a comprehensive review of NTs and attempts to identify how they enhance the model interpretability. We first propose a thorough taxonomy of NTs that expresses the gradual integration and co-evolution of NNs and DTs. Afterward, we analyze NTs in terms of their interpretability and performance, and suggest possible solutions to the remaining challenges. Finally, this survey concludes with a discussion about other considerations like conditional computation and promising directions towards this field. A list of papers reviewed in this survey, along with their corresponding codes, is available at: \href{https://github.com/zju-vipa/awesome-neural-trees}{https://github.com/zju-vipa/awesome-neural-trees}.

\end{abstract}

\ccsdesc[500]{General and reference~Surveys and overviews}
\ccsdesc[500]{Computing methodologies~Knowledge representation and reasoning}
\ccsdesc[500]{Computing methodologies~Neural networks}
\ccsdesc[500]{Computing methodologies~Classification and regression trees}

\keywords{Neural Tree, Neural Decision Tree, Interpretable Deep Learning}

\maketitle

\section{Introduction\label{sec:introduction}}
NNs and DTs are both popular machine learning models operating on broadly distinct paradigms, namely \textit{connectionist} and \textit{symbolic}, which have been compared detailedly in \cite{quinlan1994comparing}. Their primary difference is the form of knowledge learned. In general, the former learns hierarchical representations, while the latter learns hierarchical clusters. Despite their proven successes in academic and commercial communities, they usually come with mutually exclusive benefits and limitations \cite{tanno2019adaptive}. This section first presents a brief review for the pros and cons of NNs and DTs, then introduces their integrated approaches (\textit{i.e.,} NTs) that try to reach a sweet point between the two worlds.

\subsection{Review for Benefits and Limitations of NNs}
NNs are arguably the most successful machine learning models in the last decades, and their unprecedented success is primarily attributed to the hierarchical representation learning through the composition of nonlinear transformations \cite{chen2021self, zeiler2014visualizing, bengio2013deep}. Such a learning paradigm alleviates the need for feature engineering, and enables NNs to represent nonlinear relationships that are difficult to approximate through other machine learning methods. Besides, NNs do not require prior knowledge with respect to the distributions of data, allowing training to scale to large datasets with the help of stochastic optimizers. Moreover, NNs can be parallelized, appealing to tasks where rapid computation is critical. Due to the above advantages, NNs benefit from excellent performance and strong generalization capacity. However, NNs are not flawless. They suffer from several straightforward shortcomings. Firstly, \textbf{lengthy training time and expensive computation cost} are caused by the distributed representation of knowledge in the form of connections between a vast number of neurons \cite{sethi1990entropy}, so that the number of parameters to be trained can be large. Then, the \textbf{daunting architecture design process} indicates that NNs often demand a manually designed architecture so as to satisfy a specific task, requiring domain expertise \cite{zoph2016neural}. And, above all, NNs are notorious for the \textbf{lack of transparency} due to their black-box nature. It is not easy to interpret trained NNs in a physically meaningful way \cite{su2007neural}. Because the knowledge is encoded in the parameters, leading to little insight into the reasoning process of the model. This non-transparent knowledge representation is the primary obstacle to their wide application in a number of fields, such as medical diagnosis and therapy, where the model interpretability is highly demanding.

\subsection{Review of DTs and Their Developments}
\subsubsection{\textbf{Advantages of Typical DTs}}
DTs learn to partition the input space by learning hierarchical clusters of data, consisting of internal (or decision) nodes and terminal (or leaf) nodes connected by branches. An internal node contains a routing function of the type $x_j$ > $b_i$ (is feature $j$ bigger than threshold $b_i$?) for axis-aligned trees, or  $\mathbf{w_j^T}\mathbf{x}$ > $b_i$ for oblique trees, to decide which child node to visit next, while a terminal node is associated with a class label. The prediction of a DT is given by a path from the root to a leaf consisting of a sequence of tests, through which we can gain insight into the reasoning process of the model by breaking up the prediction into a sequence of intermediate and semantically meaningful decisions \cite{wan2020nbdt}. In contrast to typical NNs which need pre-determined architectures, DTs grow recursively and greedily through some heuristics based on informativeness-related criteria, namely data-driven architectures. DTs thus benefit from lightweight inference and transparent decision-making mechanism that are absent in NNs. Some of the most representative DT algorithms are CART \cite{loh2011classification}, ID3 \cite{quinlan1986induction}, C4.5 \cite{quinlan1987simplifying}, etc. A comprehensive review of DTs can be found in \cite{loh2011classification, safavian1991survey}. It is also common to use an ensemble of multiple trees, such as Random Forest \cite{breiman2001random} and XGBoost \cite{chen2016xgboost}, to boost performance at the expense of interpretability.

\begin{figure}
\centering
\includegraphics[width=\textwidth]{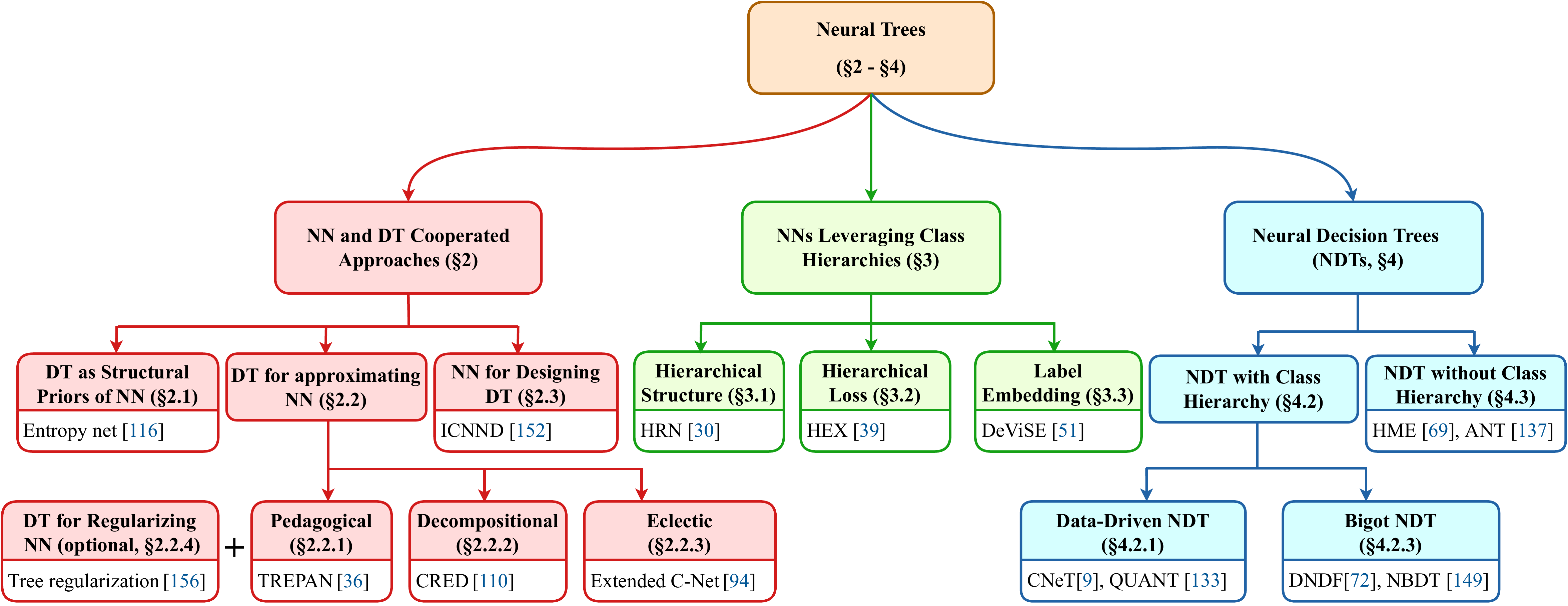}\\
\caption{The taxonomy of NTs (NNs and DTs integrated approaches) which are categorized into three main domains. The more fine-grained categorization will be discussed detailedly in later sections. Each category is demonstrated with \textbf{a part of representative works} in the figure.}
\label{fig:outline}
\end{figure}

\subsubsection{\textbf{Limitations and Improvements of DTs}}
Compared to NNs, DTs have the opposite limitations that are amplified with the development of deep learning. Typical DTs are only good at extracting comprehensible and symbolic rules, requiring tabular data or manually engineered features. Before the appearance of oblique and multivariate DTs \cite{breiman2017classification, murthy1994system, utgoff1990incremental, heath1993induction, murthy1993oc1, brodley1995multivariate}, scholars may ascribe the \textbf{limited expressivity of a single DT} to the use of simplistic routing functions, \textit{i.e.,} splitting on axis-aligned features with discontinuous indicator functions. These hard partitioning methods are greedy, heuristic, non-differentiable and hinder the use of gradient descent-based optimization. Oblique and multivariate trees are thereafter proposed to alleviate these problems. Oblique trees use linear combinations of variables to split the node and partition the space into more general polygonal, while multivariate trees may include non-linear combinations and produce curved surfaces. Although these approaches do not explore how to make the parameters differentiable, they allow the possibility of using connectionist approaches to take a stochastic decision at each internal node. Another shortcoming of DTs is that \textbf{they do not usually generalize as well as NNs}, because a node at the lower levels of a DT only receives a small fraction of the training data and tends to overfit \cite{frosst2017distilling}. This can be relieved by using fuzzy set theory and fuzzy membership functions \cite{zadeh1996fuzzy, zadeh1973outline}, where the partial memberships between nodes allow each sample to contribute to each leaf. The extension of expressive capabilities transforms DTs into a more powerful functional approximant, namely fuzzy decision trees (FDTs) — DTs utilizing fuzzy memberships for routing functions, otherwise crisp decision trees (CDTs). It incorporates features of connectionist methods and makes it possible to design globally optimal trees on which continuity constraint is imposed to improve performance. FDTs also alleviate another pitfall of CDTs, \textit{i.e.,} the \textbf{zero-tolerance of mistakes} that once a wrong path is taken at an internal node, no way of recovering from this mistake can be obtained \cite{sethi1990entropy}. \hyperref[fig:generic-NDT]{Figure 7} will depict how FDTs fix the zero-tolerance pitfall. We indicate that the improvements of multivariate DTs and FDTs are theoretical bases for techniques employing NNs to design DTs. More details can be referred to in \hyperref[sec:inference-schema]{section 4.1}.

\subsection{Neural Trees: Intergration of NNs and DTs}
As mentioned above, NNs and DTs have their own considerations, advantages and limitations. The two models progressed independently until the 1990s, then researchers began to investigate ways of combining them. During the last three decades, there has been a significant amount of work exploring the intersection of NNs and DTs. We unify them into a concept: neural trees (NTs). As illustrated in \autoref{fig:outline}, we categorize NTs into three domains, which implies the gradual integration and co-evolution between NNs and DTs.

\subsubsection{\textbf{Non-hybrid: NNs and DTs Cooperated Approaches}} These approaches are the very first to combine NNs and DTs. It can date back to the early 1990s when DTs were supposed to provide structural priors for NNs \cite{banerjee1990initializing, sethi1990entropy, sethi1990layered, ivanova1995initialization} or extract rules from a trained NN \cite{krishnan1999extracting, craven1995extracting}. They are perceived implicit combinations of NNs and DTs, as they do not bring the two worlds into one hybrid model. Since deep neural networks (DNNs) were not developed then, it may be appropriate to utilize DTs to design or approximate a NN with a small number of hidden layers, and DTs may still be competitive with NNs in many fields. However, with the rapid development of deep models in this century, typical DTs can no longer accomplish the same work on DNNs or compete with DNNs due to their limited expressivity.

\subsubsection{\textbf{Semi-hybrid: NNs leveraging class hierarchies}\label{sec:intro-class-hierarchy}} Based on the view that DTs are class hierarchies implemented by decision branches, this survey tries to include NNs that draw on \textbf{a part of} ideas from DTs, either the class hierarchy or the decision branches, \textbf{but not both}. However, we will only concern about the former in our taxonomy, \textit{i.e.,} NNs leveraging class hierarchies. These approaches can not implement the class hierarchy in the network structure due to the absence of decision branches. They instead resort to incorporating the hierarchical relations into a NN directly. As for NNs utilizing decision branches, we find their decision mechanisms are quite different from those in typical DTs (\textit{i.e.,} they are not parameter-free and informativeness-based), so we do not comprise them in the taxonomy. We instead provide a brief discussion in \hyperref[sec:half-integration]{section 5.3} and indicate their relations to special DTs whose routing functions are implemented by NNs. These approaches are considered to be half and implicit integration of NNs and DTs. They borrow some inherent ideas from DTs for NNs, instead of designing a NN that works like a DT.

\subsubsection{\textbf{Hybrid: Neural Decision Trees (NDTs)}} In contrast to the approaches mentioned above, NDTs are hybrid NN models that implement both the class hierarchy and the decision branches. The earliest NDTs were proposed around 1990, when scholars trained DTs to mimic the input-output pattern of a NN \cite{jordan1994hierarchical, stromberg1991neural} and apply them to handle low-dimensional tabular data. With the progress of DNNs, recent NDTs have scaled up to datasets like CIFAR10 \cite{krizhevsky2009learning}, CIFAR100 \cite{krizhevsky2009learning}, ImageNet \cite{deng2009imagenet}, etc. However, these models tend to be more interpretable at the cost of performance. In order to make up for this, many trade-offs are adopted, \textit{e.g.,} feeding the NDT with latent representations to meet its preference for tabular data \cite{ji2020attention, kontschieder2015deep, wan2020nbdt, nauta2021neural}. In \autoref{sec:NDT} and \hyperref[sec:ndt-trade-off]{section 5.2} we will demonstrate that such trade-offs between performance and interpretability are quite common in NDTs.

\subsection{Outline of This Survey}
This survey proposes a taxonomy of NTs (\textit{c.f.,} \autoref{fig:outline}), and the more fine-grained categorization will be described in \autoref{sec:DT_Assist_NN}, \autoref{sec:NN_with_hierarchy} and \autoref{sec:NDT}. Meanwhile, the pros and cons of different NTs will be discussed. Afterward, we provide analysis for these approaches in \autoref{sec:Analysis_Comparison} with regard to their interpretability and performance, and propose possible solutions for the remaining challenges among them. Finally, other considerations and promising directions in this field will be discussed, followed by a brief conclusion in \autoref{sec:conclusion}. For the sake of coherence and rationality, this survey will concern most about how these approaches allow interpretation in NNs by means of DTs. Therefore, our introduction and analysis will focus on NDTs that are built to be inherently interpretable, followed by other NTs related closely to the model interpretability.

\section{Non-hybrid: NNs and DTs Cooperated Approaches\label{sec:DT_Assist_NN}}
This section introduces NTs employing DTs as auxiliaries for NNs as well as using NNs as tools to improve the design of DTs. In these approaches, "neural" and "tree" are separated, \textit{i.e.,} one of NN and DT is assigned to accomplish a specific task, while the other performs as its assistant or interpreter. NNs and DTs still operate on their own paradigms and no hybrid model is produced.

\subsection{DTs as Structural Priors of NNs\label{sec:DT-structural-prior}}

\begin{figure}
\centering
\includegraphics[width=\textwidth]{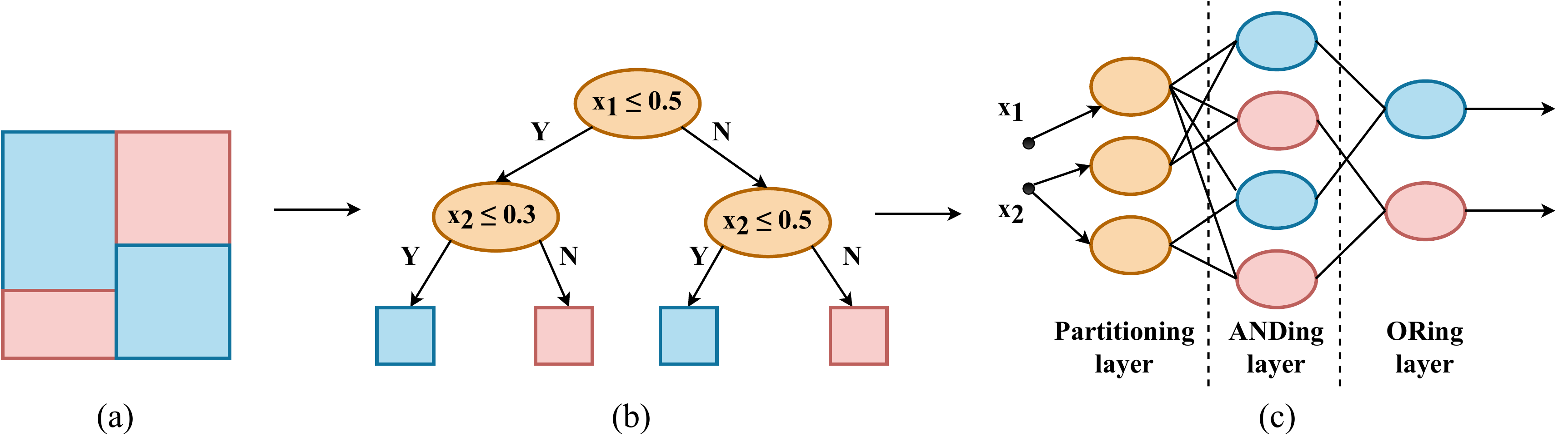}\\
\caption{An example for the entropy net \cite{sethi1990entropy, sethi1990layered, sethi1990comparison} (a) Decision boundaries of a simple binary classification problem \cite{sethi1990entropy}. (b) An axis-aligned DT designed to solve the problem in (a).  (c) The three-layered network derived from the DT, with the mapping rules proposed in entropy net \cite{sethi1990entropy}.}
\label{fig:structural_prior}
\end{figure}

The performance of NNs is believed to be sensitive to the initial weights and network architectures. Before the development of deep models, this usually means the number of hidden layers and the neurons in them. This shortcoming can be alleviated if some approximation of the target concept in terms of a logical description is available \cite{ivanova1995initialization}. Therefore, it is possible to adopt DTs to generate such logical descriptions and incorporate them in NNs. This is the direction Sethi \textit{et al.} take first \cite{sethi1990entropy, sethi1990layered, sethi1990comparison}. They propose a method to implement a pre-determined DT via a multilayer feedforward NN, termed as \textit{entropy net}. The core idea is to propose certain rules so as to make a DT and a layered NN equivalent in terms of the input-output mapping. They first construct a DT using the AMIG algorithm \cite{sethi1982hierarchical}. Subsequently, they map the DT into a three-layered NN. As shown in \autoref{fig:structural_prior}, there are natural correspondences between the DT and the mapped NN. Neurons in the first layer correspond directly to internal nodes of the DT. In the original paper, the DTs are designed as axis-aligned trees, so a certain neuron in the first layer evaluates one axis-aligned hyperplane test. This layer is called the \textit{partitioning} layer. Similarly, each neuron in the second hidden layer corresponds to a leaf node, and connections to this neuron correspond to the path from the root to this leaf. Since the conditions along any particular path must be satisfied to reach a particular leaf, the second layer is termed as \textit{ANDing} layer, where each neuron implements an \textit{AND} operation on a set of half spaces. Finally, the neurons in the output layer correspond to distinct classes. Paths that lead to the same class will be connected to the same neuron. The output layer is thus dubbed as the \textit{ORing} layer, aiming to collect the closed decision regions associated with specific classes. After the entropy network is constructed, they associate the sigmoid or other soft nonlinearities with every neuron and train the network except the first layer.

The entropy net relieves empirical means in the design of MLP networks and permits interpretation of the knowledge embedded in the generated connections and weights. The resultant network usually has fewer connections, and is endowed with the fault-tolerance ability that is absent in the original DTs. However, the DT-inducing process does not change. They still need to determine appropriate features and decision rules for the routing functions, and still use a single feature to split (\textit{i.e.,} do not conduct on multivariate trees), so it is possible to end up with huge trees. Moreover, their design can only map a DT into a three-layered NN, limiting its performance and scalability.

Several works are developed from or inspired by the entropy net. Cios \textit{et al}. \cite{cios1992machine} propose an ID3-based algorithm that converts DTs into hidden layers until the learning task becomes linearly separable at the output layer. Sethi \cite{sethi1995neural} incorporates soft thresholding in the mapped NN by adjusting the weights of connections or the slopes of the output functions. In \cite{park1994mapping} and \cite{brent1991fast}, authors propose methods for mapping oblique trees to NNs. In this case, each internal node of the tree evaluates an oblique hyperplane test, so does each neuron in the first layer of the mapped NN. Compared with entropy nets, the use of oblique trees instead of axis-aligned trees generalizes the DT—NN mapping in terms of the number of features, thereby providing heuristic but more efficient rules and determining a more appropriate size for the mapped NN. Particularly, Tsujino \textit{et al}. \cite{tsujino1995implementation} develop a sophisticated knowledge acquisition system that cares about the knowledge form DTs and NNs prefer. They induce the \textit{knowledge structure} in the form of a DT from symbolic examples, whereas refine the restructured NN with numerical examples. Furthermore, a few approaches \cite{ivanova1995initialization, banerjee1990initializing, setiono1999mapping} decide to traverse the DT to create a disjunctive normal form formula for each class, where the atoms are attribute-value tests such as $x_{1} < 0.5$. Afterward, they redescribe the formulas and the training examples in terms of interval-membership functions, and construct an entropy net replacing the partitioning layer with their interval layer.

These approaches aim to derive NNs from DTs, so the functions of an entropy network-like system can be significantly affected by the tree-induction method and the size of the training set. With the progress of DNNs, many excellent methods have emerged for determining the network structure or initializing its weights. As a result, DT-derived NNs are no longer attractive because it is hard to map a DT into a DNN with many layers and connections. However, their ideas of employing DTs to assist the design of NNs remain instructive.

\subsection{DTs for Approximating NNs \label{sec:DT-interpreting-NN}}
For many learning tasks, it is important to obtain classifiers that permit both the high performance and the human-intelligible interpretation. NNs are limited in this respect since they are usually difficult to interpret after training. We deem that the performance of NNs may frequently reach a bottleneck, but the model interpretability is far from satisfactory. We can gain little insight into the causes and effects of a feature activation in a meaningful way. This section focuses on using DTs to interpret a trained NN, based on the view that solutions formed by symbolic systems are much more amenable to human comprehension \cite{craven1995extracting}. These algorithms perform \textit{post-hoc} interpretation by adding the explainability objective after training so as to approximate the network and mimic the decision boundaries implicitly learned by the hidden layers. In some literature \cite{andrews1995survey, craven1994using, schmitz1999ann}, these methods are summarized as "rule extraction" techniques that extract comprehensible and symbolic rules from NNs, and can be further categorized by the approach used to characterize the internal model of the network: \textit{pedagogical} (black-box rule extraction), \textit{decompositional} (link rule extraction) and  \textit{eclectic} (mixture of both). In this survey, we only consider those related to DTs.

\subsubsection{\textbf{Pedagogical Techniques}} 
\begin{figure}
\centering
\includegraphics[width=\textwidth]{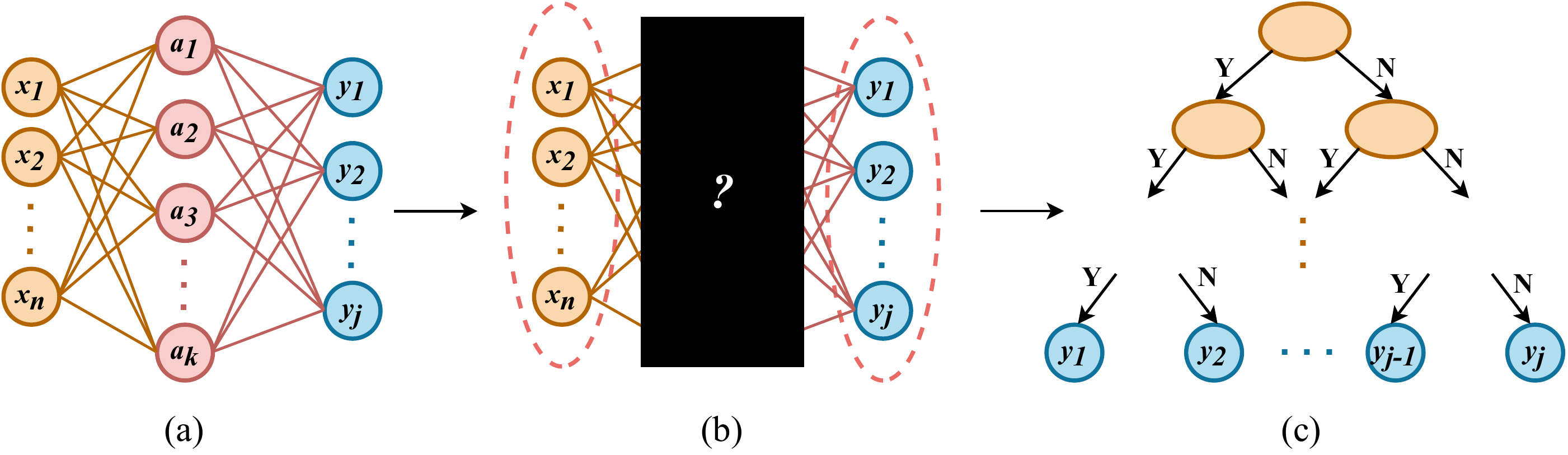}\\
\caption{A description of pedagogical techniques. (a) A NN with a single hidden layer needs to be interpreted. (b) The black-box approach in view of pedagogical techniques, from which only the activation value of input and output units can be obtained. (c) The DT derived from (b), where the ellipse nodes indicate that the tests may not be axis-aligned.}
\label{fig:rule_extraction}
\end{figure}

They treat NNs as black-box approaches, and extract the input-output rule directly form NNs regardless of its intermediate layers \cite{thrun1993extracting, thrun1994extracting, craven1994using, narazaki1996reorganizing}. Carven \textit{et al.} \cite{craven1994using, craven1995extracting} are the very first to propose such an algorithm based on DTs, named \textit{TREPAN} (\textit{cf.,} \autoref{fig:rule_extraction}). This algorithm aims to label examples as predicted by the trained network, so that a DT can be induced to approximate the concepts represented by the NN. After labeling training examples, they use the ID2-of-3 algorithm \cite{murphy1991id2} to form trees. At a given node to be expanded, they will call the trained NN to evaluate a set of candidate splits. Then choose the one with greatest potential to increase the \textit{fidelity}, defined as the percentage of test examples on which the prediction made by a DT agrees with its NN counterpart. The trained network is also assigned to generate artificial examples when adequate training data is not available. It enables the DT to use as many instances as desired to select each split, and is in a sense taking advantage of NNs like robustness to noise, stable learning and better generalization \cite{krishnan1999extracting}. 

\textit{TREPAN} is proposed at a time of lacking general methods for understanding the knowledge encoded in the trained network. It is a promising advance towards this direction and inspires people to explore knowledge by querying and sampling NNs. Nevertheless, the original algorithm is limited to classification tasks and incapable of building DTs from models that were continuous in nature \cite{young2011investigation}. The use of the $m$-of-$n$ type splits also complicates its applications for high-dimensional problems. Therefore, a number of approaches are developed to improve and generalize \textit{TREPAN}.

On the one hand, some approaches focus on the improvement during the DT-induction process. In \cite{vasilev2020decision}, the authors remove all insignificant input neurons from the trained network before inducing the DT so as to decrease the number of rules. Several methods aim to turn the $m$-of-$n$ type splits into standard splitting tests \cite{dancey2004decision, boz2002extracting, boz2000converting, krishnan1999extracting, zhou2004nec4}. Dancey \textit{et al}. \cite{dancey2004decision} use the same sampling methods as that in \textit{TREPAN}, but construct a CART tree with the sample points. Boz \textit{et al.} \cite{boz2002extracting, boz2000converting} propose to extract C4.5-like DTs with a new discretization algorithm to handle continuous features, and provide four splitting techniques to increase fidelity. Zhou \textit{et al}. \cite{zhou2004nec4} likewise employ C4.5 algorithm, but they use NN ensemble instead of a single NN to derive DTs. Krishnan \textit{et al}. \cite{krishnan1999extracting} adapt a genetic algorithm-based method \cite{eberhart1992role} to generate artificial examples and perform selection algorithms \cite{specht1990probabilistic, kohonen2012self} to filter outliers, then an arbitrary induction algorithm can be applied to construct DTs. 

On the other hand, some approaches concern more about generalizing \textit{TREPAN} algorithm to other situations or applications. Young \textit{et al.} \cite{young2011investigation} modify the implementation of \textit{TREPAN} in order to develop and test DTs derived from continuous-based models. Rangwala \textit{et al}. \cite{rangwala2004trepan} propose several heuristics to improve the performance of \textit{TREPAN}, and extend it to multi-class regression problems. Faifer \textit{et al}. \cite{faifer1999extracting} develop \textit{TREPAN} in two ways: use fuzzy representation during the tree-induction process, and use additional heuristic approaches for generating artificial examples. More recently, Muller \textit{et al.} \cite{muller2022dt+} resort to incorporating DTs into graph neural networks (GNNs). They first create the \textit{Diff-DT+GNN} layer, where the nodes update their state through MLPs taking the aggregated messages from neighbors and the node's current state as input. Afterward, they use the Diff-DT+GNN layers to query all the training graphs and obtain results, through which DTs are trained to replace each block predicting the outputs from inputs.

While most approaches above evaluate each split by a kind of fidelity test of the extracted tree to the network, Schmitz \textit{et al.} \cite{schmitz1999ann, aldrich1997characterization} provide a special attribute selection criterion for inducing axis-aligned trees. Their algorithm, termed as \textit{ANN-DT}, examines the responses of the trained NN in the feature space and conducts a significance analysis of different attributes pertaining to these responses. Considering a NN which models the function $f$, they define the \textit{absolute variation} $v(f)$ of the function $f(x)$ between two sample points $i$ and $j$ as the absolute value of the directional derivative of $f(x)$ integrated along a straight line between the two points, \textit{i.e.,}
\begin{equation}
    v_{ij}(f) = \int_{x_{j}}^{x_{i}} \left | \bigtriangleup f(x) \cdot u \right | dx,
\end{equation}
where $\mathbf{u}$ is the unity vector in direction $\mathbf{x_{i}} - \mathbf{x_{j}}$. The \textit{significance} is thereafter measured by the correlation between the absolute variation of that attribute taken from all possible points in the NN-sampled data set. At a given node, the attribute with the maximum significance was selected, and the threshold for splitting is chosen by minimizing the weighted variance.

The algorithms of pedagogical techniques for NN-approximation have a profound impact on DT-based interpretations. However, they use only the activation value of input and output units to extract rules. The only interface permitted with the NN is presenting an input and obtaining the output \cite{krishnan1999extracting}, which greatly limits the inclusion of the knowledge present in the intermediate layers.

\subsubsection{\textbf{Decompositional Techniques}}
In contrast to pedagogical techniques that build DTs only based on the input-output mapping of trained NNs, decompositional strategies also concern units in the hidden layers \cite{towell1993extracting, fu1991rule, gallant1993neural}. They extract rules from the trained NN at the level of individual neurons, thus gaining insight into their inner structures.

Some decompositional techniques of utilizing DTs aim to deal with NNs with one hidden layer and often produce axis-aligned trees \cite{sato2001rule, zilke2016deepred, al2012eclectic}. Sato \textit{et al.} \cite{sato2001rule} present the CRED algorithm. It discretizes the activation values of each hidden unit, and use the C4.5 algorithm to induce a hidden-output tree. By this mean intermediate rules like $(a_{1} > 0.5) \wedge (a_{2} > 0.6) \Rightarrow y_{1}$ are extracted, where $a_{i}$ is the activation value of the hidden neuron $i$. For each term used in these tests, another DT is induced to extract rules describing the state of the hidden neurons in terms of the input variables, \textit{e.g.,} $(x_{1} \le 0.5) \wedge (x_{2} = 0.3) \Rightarrow a_{1} > 0.5$. Finally, rules from two steps are substituted and merged into a DT describing rules that entail the class variable to the input atomic terms. Zilke \textit{et al}. \cite{zilke2016deepred} extend this algorithm by deriving additional DTs and intermediate rules for every additional hidden layer, so that it can be applied to deeper NNs. However, such a successive substitution may result in incomprehensible rules with many redundancies. Possible solutions can be found in \cite{al2012eclectic}, another decompositional approach where rules are simplified by logical simplification \cite{brayton1984logic}.

However, axis-aligned trees derived by these algorithms may lead to an intractably large number of rules, and tend to be improper because the decision boundaries of a NN are usually not parallel. Consequently, some decompositional approaches design algorithms for extracting oblique and multivariate trees \cite{setiono1997neurolinear, nguyen2020towards}, where the rules include combinations of multiple attributes. Setiono \textit{et al}. \cite{setiono1997neurolinear} extract oblique decision rules from NNs. Each condition of a rule corresponds to a hyperplane that is not necessarily axis-aligned, which reduces the number of rules. More recently, Nguyen \textit{et al.} \cite{nguyen2020towards} introduce two novel multivariate DT algorithms for rule extraction, named the Exact Convertible Decision Tree (EC-DT) and the Extended C-Net algorithm. The former adopts the decompositional strategy to layer-wise extract rules that represent the exact decision boundaries learned by the hidden layers, while the latter is an eclectic method combining the decompositional approaches from EC-DT with a pedagogical tree learning algorithm to extract rules.

\subsubsection{\textbf{Eclectic Techniques}} They combine elements of pedagogical and decompositional techniques. The Extended C-Net algorithm \cite{nguyen2020towards} mentioned above is a typical eclectic technique. It derives an axis-aligned tree from the relationship between the last hidden layer and the outputs, then infers the input-output relationship through the recursive back-projections guided by the weights of the NN. In the previous literature, eclectic techniques usually refer to those drawing inferences from the magnitudes of the weights in a NN \cite{tickle1994dedec, sestito1992automated}. However, few of them are based on DTs. Therefore, we also include approaches aiming to extract \textbf{partial knowledge} contained in the hidden layers and do not demand their approximation target is the input-output relationship \cite{yang2018global, zhang2019interpreting}. Yang \textit{et al}. \cite{yang2018global} design a CART tree learned from the contribution matrix consisting of the contributions of input variables to predicted scores for each single prediction. In order to apply this interpretation tree to the high-dimensional scene understanding tasks, they adopt a scene parsing algorithm and dilated convolutional network to segment each image into semantically meaningful parts, from which the contribution matrix can be calculated. Recently, Zhang \textit{et al}. \cite{zhang2019interpreting} propose to learn a DT that encodes all potential decision modes of the convolutional neural network (CNN) in a coarse-to-fine manner by decomposing feature representations in high conv-layers of the CNN into elementary concepts of object parts. This interpretation tree can tell people which object parts activate which filters and how much each object part contributes to the prediction score. We perceive these approaches as generalized eclectic techniques, as they do not simply approximate the input-output relationship, yet do not perform explanation at the level of individual neurons. They instead partially opened the black box with varied strategies.

\subsubsection{\textbf{Optional: DTs for Regularizing NNs} \label{sec:DT-regularization}}
Despite the above success of DT-based NN approximation, simply fitting a DT to a trained NN without regularization usually leads to unsatisfactory results in terms of accuracy and fidelity \cite{schaaf2019enhancing}, so it is beneficial to train NNs that resemble compact DTs ahead. Wu \textit{et al}. \cite{wu2018beyond} propose a complexity penalty function named \textit{tree regularization}, which aims to optimize the deep model for interpretability and human-simulatability. The tree regularization favors models whose decision boundaries can be well-approximated by small DTs and encourages the deep model to resemble an axis-aligned DT in pre-defined, human-interpretable contexts. They further extend their work \cite{wu2021optimizing, wu2020regional} to address the unreasonable expectation for a single DT to predict well across all possible inputs. Their reformed algorithm encourages a deep model to be approximated by several separate DTs specific to pre-defined regions of the input space. However, the DT needs to be determined before the NN so as to calculate the regularization loss, which is not differentiable. They made up for this by using an additional surrogate network to estimate the loss, but this probably results in lengthy training time and cumbersome tuning. Schaaf \textit{et al}. \cite{schaaf2019enhancing} address this by using L1-orthogonal regularization during training.

These approaches provide an auxiliary step before executing the rule extraction algorithms introduced before. They try to influence the nonlinear decision boundaries of the network and allow it to be well-approximated by small DTs. But every coin has two sides. These methods somehow lead to a few compromises in accuracy because they do not allow the NN to produce arbitrary decision boundaries anymore, and they do not make a remarkable contribution to the model interpretability as they are usually designed to serve the post-hoc approximating approaches.

\subsubsection{\textbf{Summary}\label{sec:summary-approximation}}
There are three lines of techniques, among which the decompositional approaches are superior in preserving the structure and fidelity, whereas other approaches may generate more compact and effective trees \cite{zhang2019interpreting}. For performance, these approaches can not induce a DT whose performance matches its NN counterpart. Because the DT was distilled from the NN, which acts as an \textit{oracle} for obtaining the indisputable result corresponding to any input presented to it. In this case, these algorithms may offer excessive tolerance for the mistakes NNs may make. Besides, it can be a challenge to apply these approaches to DNNs, because a rule-based model with limited expressivity can hardly approximate a DNN. For interpretability, these methods all belong to post-hot analysis. They assign a DT to perform interpretation owing to its natural advantages in reasoning and attribution. However, these techniques aim to understand already learned models by fitting explanations, rather than design models that are inherently interpretable. A NN capable of generalizing underlying trends in the data has to be constructed first.
Moreover, it is well-known that black-box models often have multiple optima of similar predictive accuracy \cite{goodfellow2016deep}, thus making the simulation of a post-hot interpretation untrustworthy. It can not explain the reasoning process of how the network \textit{actually} makes its decisions.

\subsection{NNs for Designing DTs\label{sec:NN_design_DT}}
These approaches in fact comprise all the DT-based rule extraction methods introduced above. When they help NNs with interpretation, NNs also provide them with prior knowledge and endow them with generalization ability. Therefore, there is mutual support between the NN and the DT in these approaches. However, this section will introduce a particular algorithm proposed by Wang \textit{et al.} \cite{wang2005integrated}, where NNs are specialized for designing more reasonable DTs and do not benefit from it. Their core idea is simple: induce DTs from more valid data filtered by the trained NNs. They use \textit{sensitivity} to denote the different contributions of the input variables, calculated by the difference between predictions from the NN when the variable is removed and when it is left in place. In this way, the irrelevant variables will be discovered and eliminated. Afterward, they adopt the NN to examine all examples in which the noise data will be removed. Finally, they will condense the training set again by clustering and induce a C4.5 tree with the worked dataset.

Although it is certainly feasible to specialize NNs to design DTs, it is rare in practice and this approach is the only instance to our knowledge. This is probably because NNs are considered to be much more powerful machine learning models, which are too prodigal to serve DTs. This approach looks close to pedagogical techniques in NN-approximation, but it differs in that it does not apply the trained NN to label examples or generate artificial examples, so it does not perform any model approximation. It should also be noted that their \textit{sensitivity} is different from the \textit{significance} in ANN-DT algorithm in both calculation and application, but the idea of adopting network analysis usually shows appreciable advantages over greedy attribute selection in typical DTs. This is somewhat a kind of knowledge distillation where the NN performs as the teacher telling its DT student "what inputs matter the most for which the trained NN gives a desired output."

\section{Semi-hybrid: NNs Leveraging Class Hierarchies\label{sec:NN_with_hierarchy}}
DT algorithms have two key ideas: \textbf{decision} and \textbf{tree}. We view that "decision" is decision branches implemented by routing functions, and "tree" is the model topology that identifies a class hierarchy. Therefore, our NT-methods includes NNs that draw on a part of ideas from DTs. As introduced in \hyperref[sec:intro-class-hierarchy]{section 1.4.3}, we only concern NNs leveraging class hierarchies in this section, while NNs utilizing decision branches will be discussed later in \hyperref[sec:half-integration]{section 5.3}. In these approaches, the class hierarchies are present in the form of a tree, named \textit{label tree}. Each node of the tree will be associated with a class label, and the connections between the parent and its children nodes represent the inclusion relationships of classes in different grains. The labels of the tree can be organized according to domain knowledge or automatically generated by algorithms \cite{wang2021label}. In our view, a DT directly implements a label tree where the classes progress from generic to specific along the tree, \textit{i.e.,} the leaf nodes represent classes at the finest grain and each internal node corresponds to a superclass at a certain grain that contains all subclasses in its descendant nodes. In this case, each node will correspond to a concept that can be either actual or abstract, (\textit{i.e.,} the super classes do not have to be semantically meaningful). On the contrary, methods of exploiting the class hierarchy in NNs have to implicitly encode the label tree into the networks due to the absence of decision branches. Some literature \cite{wang2021label, bertinetto2020making} indicate that there are three lines of research: hierarchical architecture, hierarchical loss function and label embedding based methods, corresponding to different strategies of employing class hierarchies. Particularly, Bertinetto \textit{et al}. \cite{bertinetto2020making} provide a framework for better understanding of these approaches. Considering a training set $S = \left \{ \left ( x_{i}, C_{i} \right )  \right \} _{i=1,\dots N}$ and a flat network modeling the predictor function $\phi (x;\theta )$, the parameters $\theta$ are usually learned by minimising
\begin{equation}
    \frac{1}{N}\sum_{i=1}^{N}\mathcal{L}( \phi (x_{i};\theta),y(C_{i})) + \mathcal{R}(\theta ),
\end{equation}

\begin{figure}
\centering
\includegraphics[width=\textwidth]{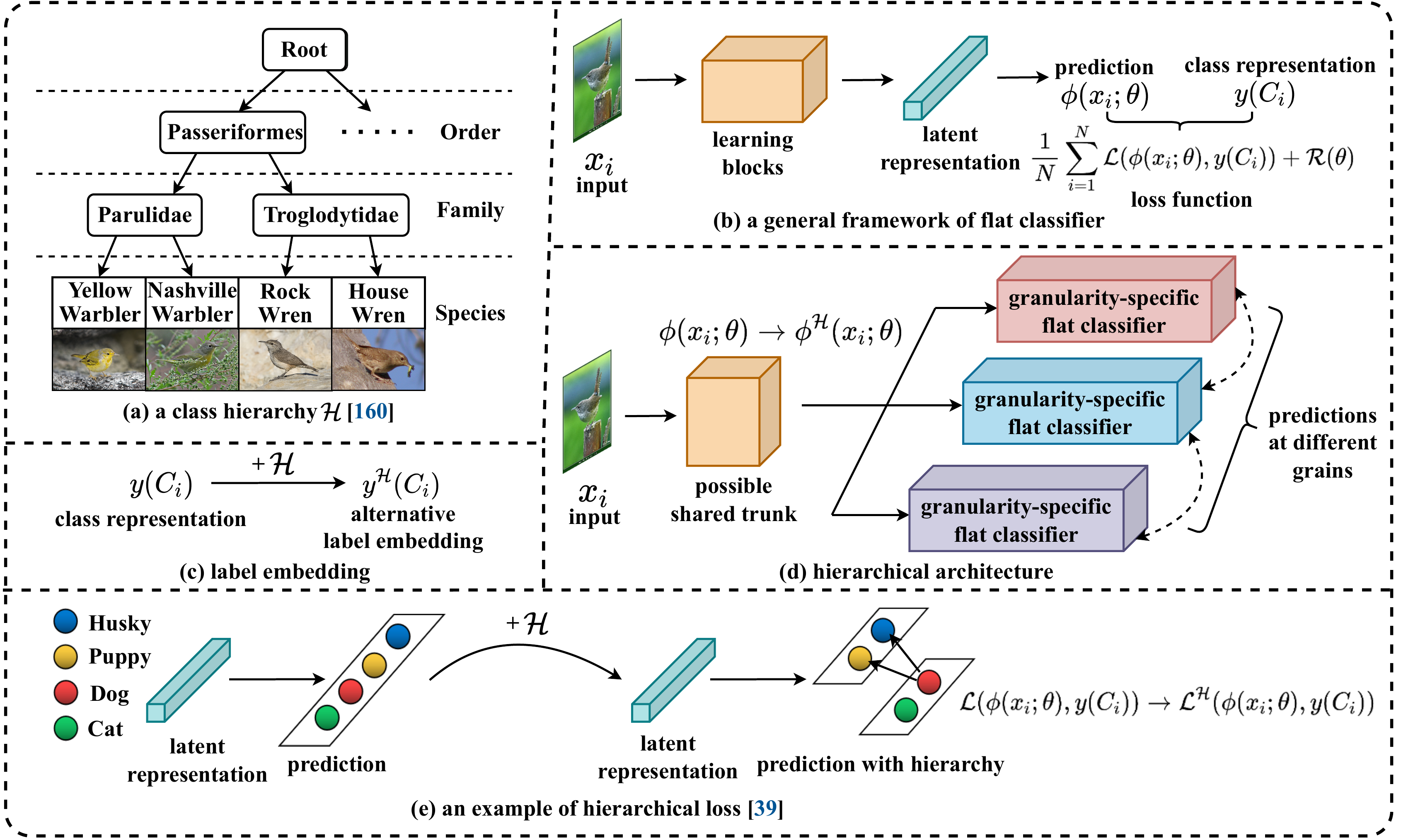}\\
\caption{Overview of NNs leveraging class hierarchies. (a) An exemplary class hierarchy $\mathcal{H}$ \cite{wang2021label}. (b) A generic framework of flat classifiers, where the loss function $\mathcal{L}$ \cite{bertinetto2020making} needs the participation of $\mathcal{H}$. (c) Encode $\mathcal{H}$ into the label embeddings. (d) Hierarchical architectures with each branch responsible for a certain grain. The dashed lines represent possible interactions. (e) An example of incorporating $\mathcal{H}$ into loss functions \cite{deng2014large}.}
\label{fig:class_hierarchy}
\end{figure}

\noindent where $y(C_{i})$ is the embedded representation of the class label $C_{i}$ and the loss function $\mathcal{L}$ is tasked to compare it with the output of the predictor $\phi(x_{i};\theta)$. $\mathcal{R}$ is the regulariser. By now the framework is still agnostic of the class hierarchy $\mathcal{H}$, so we should incorporate $\mathcal{H}$ into the loss function $\mathcal{L}$, \textit{i.e.,} hierarchical architecture ($\phi(x_{i};\theta) \to \phi^{\mathcal{H}}(x_{i};\theta)$), hierarchical loss function ($\mathcal{L}(\phi (x_{i};\theta),y(C_{i})) \to \mathcal{L}^{\mathcal{H}}(\phi (x_{i};\theta),y(C_{i}))$) or label embedding ($y(C_{i}) \to y^{\mathcal{H}}(C_{i})$). Details are depicted in \autoref{fig:class_hierarchy}.

\subsection{\textbf{Hierarchical Architecture}}
These methods attempt to incorporate $\mathcal{H}$ into the architecture of the classifier, so that the networks are designed to be branched and each branch is tasked to identify the concept abstraction at one specific level of the class hierarchy. Cerri \textit{et al.} \cite{cerri2014hierarchical, cerri2016reduction} propose to incrementally train a MLP for each level. Predictions made by an MLP at a given level are also provided as inputs to the MLP in the next level, by which the feature vector of the instance is augmented. Yan \textit{et al}. \cite{yan2015hd} propose to embed deep CNNs into a two-level class hierarchy. It separates easy classes using a coarse classifier while distinguishing difficult classes by the fine classifiers. We also notice that many approaches prefer to branch the network after a shared trunk \cite{wu2016learning, bilal2017convolutional, chen2018fine}. In \cite{wu2016learning}, a single network backbone is shared by multiple fully-connected layers, with each one responsible for the label prediction at its level. Alsallakh \textit{et al}. \cite{bilal2017convolutional} use a deep CNN to fit the finest-grained labels and add branches at the intermediate layers to fit the coarser-grained labels. Chen \textit{et al}. \cite{chen2018fine} likewise employ a multi-head architecture to output the prediction in a level-wisely manner, meanwhile, an attention mechanism is introduced to incorporate the prediction of coarse-grained results to guide learning finer-grained features. However, chang \textit{et al}. \cite{chang2021your} discover that only fine features will better the learning of coarser classifiers, whereas coarse predictions instead exacerbate the finer-grained feature learning. They in turn disentangle coarse and fine features and only allow the finer ones to participate in coarser-grained predictions. Furthermore, several recent approaches may concern more about the interactions between different branches \cite{wang2021label, chen2022label}. Wang \textit{et al}. \cite{wang2021label} propose to learn the \textit{label hierarchy transition matrices} (LHT) whose column vectors represent the conditional label distributions of classes between two adjacent hierarchies and are capable of encoding the correlation embedded in $\mathcal{H}$. Similarly, Chen \textit{et al}. \cite{chen2022label} propose the \textit{hierarchical residual network} (HRN) in which granularity-specific features from parent levels act as residual connections and are added to features in children levels.

It is noteworthy that such hierarchical architectures are quite different from those in typical DTs, \textit{i.e.,} \textbf{there are no decision branches}. The network of each branch is a \textit{flat classifier}. It receives the whole training data and directly predicts at its own grain, yet a typical DT only assigns partial data (hierarchical clusters) to each node and predicts only when an example reaches the leaf. This is a difference between this survey and some other literature \cite{bertinetto2020making, wang2021label, sethi1990entropy, jordan1994hierarchical} where approaches similar to DTs also belong to hierarchical architectures.

\subsection{\textbf{Hierarchical Loss Function}}
A range of literature resorts to incorporating $\mathcal{H}$ in the loss function. That is, the network is without a hierarchical structure, but the loss function exploits the underlying label tree so as to produce predictions coherent with $\mathcal{H}$. Some of them use the network to predict at each grain separately, \textit{i.e.,} first take a model pre-trained on one level, then tune it using labels from other levels \cite{giunchiglia2020coherent, peterson2018learning}. These methods, together with those in hierarchical architectures, share the same strategy of introducing inductive bias. They induce certain rules when training at a grain so as to make certain constraints on other grains. On the contrary, other methods do not have to give predictions for each grain \cite{verma2012learning, srivastava2013discriminative, deng2014large, bertinetto2020making}. They instead utilize the hierarchical constraints directly so as to enjoying lightweight inference. Verman \textit{et al}. \cite{verma2012learning} provide a probabilistic nearest-neighbor classification based framework for learning a set of hierarchical metrics that reflects the underlying class taxonomy. In \cite{srivastava2013discriminative}, a DNN with priors over the parameters of the classification layer is proposed. The priors are derived from $\mathcal{H}$, in which we can discover similar classes and transfer knowledge among them. Deng \textit{et al}. \cite{deng2014large} propose the Hierarchy and Exclusion (HEX) graphs, which encodes the semantic relations into a directed acyclic graph (DAG) and compute a loss defined on it. As shown in \autoref{fig:class_hierarchy}, they replace the traditional classifiers with their model so as to output probabilities coherent with the constraints on $\mathcal{H}$. More recently, Bertinetto \textit{et al}. \cite{bertinetto2020making} propose to incorporate $\mathcal{H}$ into the cross-entropy loss. They factorize the output of the classifier in terms of the conditional probabilities along the path from the root to the leaf of the label tree, then define the loss as the weighted sum of the cross-entropies of these conditional probabilities.

In these approaches, the loss function itself is parametrized to encourage the network to output probabilities consistent with the pre-defined $\mathcal{H}$. Predictions of more distant relatives of the target class will result in a higher penalty \cite{bertinetto2020making}. Compared to NNs with hierarchical architectures, these methods enjoy fewer parameters and more efficient computing as they allow a flat network to accomplish tasks demanding the support of a class hierarchy.

\subsection{\textbf{Label Embedding}\label{sec:label-embedding}}
The last direction for exploiting $\mathcal{H}$ is label embedding-based methods. They aim to encode $\mathcal{H}$ into embeddings whose relative locations or possible interactions represent the semantic relationships \cite{bertinetto2020making}. Frome \textit{et al}. \cite{frome2013devise} propose the DeViSE method that identifies visual objects using both labeled image data as well as semantic information gleaned from unannotated Wikipedia text \cite{mikolov2013efficient}. They use a linear mapping from visual features to embedded labels, and then a rank loss is adopted to penalize examples that are more similar to false embeddings. Bertinetto \textit{et al}. \cite{bertinetto2020making} soften the one-hot labels according to the label tree-based hierarchical factorization and calculate a loss on it. Moreover, Barz and Denzler \cite{barz2019hierarchy} map images into embeddings whose pair-wise dot products correspond to a measure of semantic similarity between classes. Dhall \textit{et al}. \cite{dhall2020hierarchical} particularly employ the entailment cones to learn order-preserving embeddings. We also notice that some approaches appeal to zero-shot classification \cite{xian2016latent, romera2015embarrassingly, akata2015evaluation}. For example, Xian \textit{et al}. \cite{xian2016latent} and Akata \textit{et al}. \cite{akata2015evaluation} similarly augment the state-of-the-art model by incorporating latent variables. They present a general framework where a compatibility function between image and class embeddings is proposed, so that matching embeddings are assigned a higher score than mismatching ones. Zero-shot classification is carried out by finding the label yielding the highest joint compatibility score.

\subsection{\textbf{Summary}}
These approaches exploit the correlations among categories of the class hierarchy, but do not implement the class hierarchy in their architectures (like a DT) due to the absence of decision branches. They mainly benefit from: (1) the cooperation between extra semantic information of the class hierarchy and the original concepts can boost the overall performance of deep models and (2) mitigate the severity of prediction mistakes, because the incorrectly classified examples may fall within semantically related categories. However, because the lack of decision branches, these approaches rarely perform stepwise inference and contribute little to the model interpretability.

\section{Hybrid: Neural Decision Trees\label{sec:NDT}}

\begin{figure}
\centering
\includegraphics[width=\textwidth]{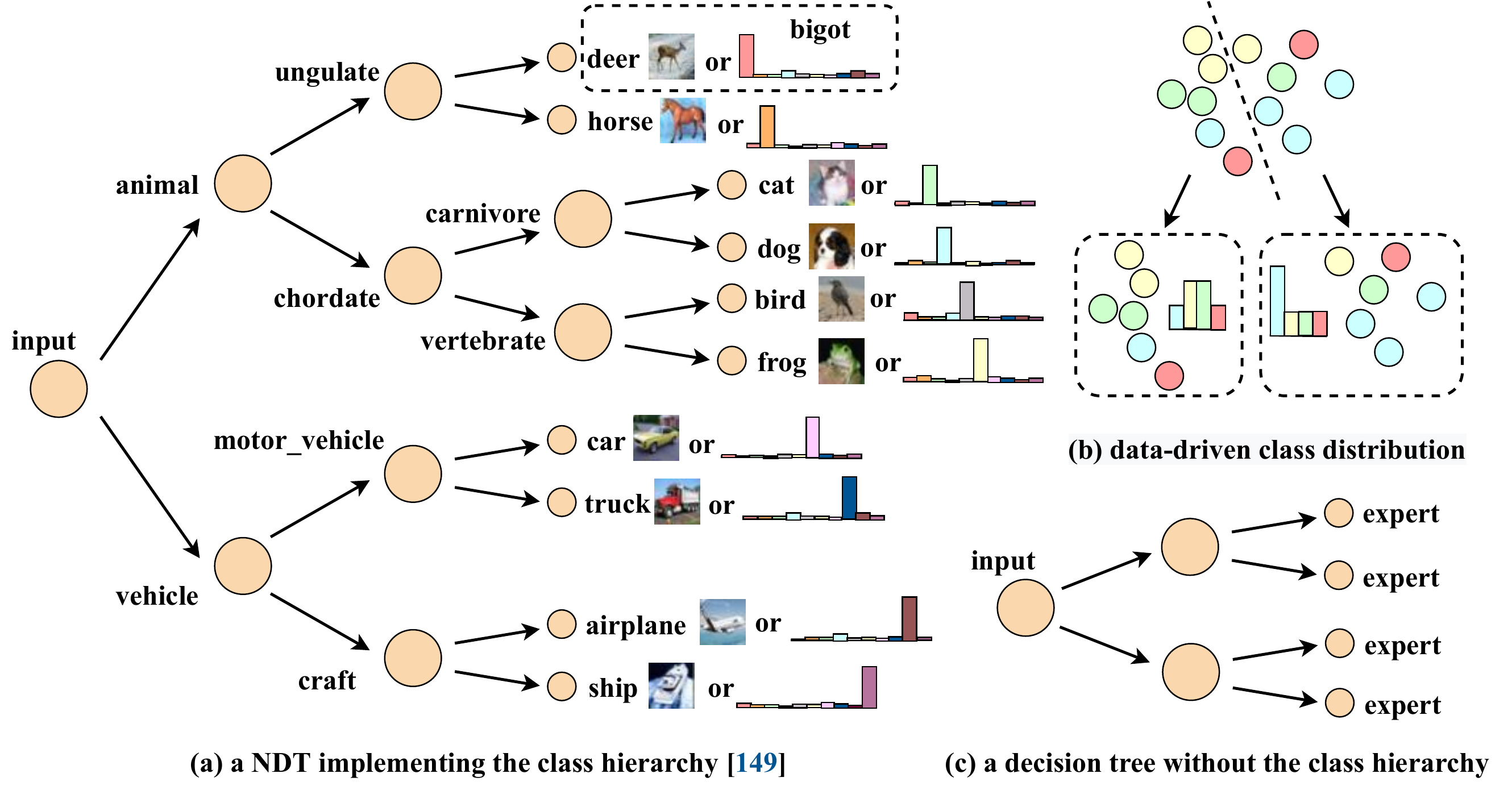}\\
\caption{A comparison between NDTs according to whether it implements a class hierarchy. (a) A NDT implementing the class hierarchy of Cifar10 dataset and induced from WordNet, which is adopted in \cite{wan2020nbdt} to relate classes. It is mainly characterised with bigot leaves, which represent determined classes or probability distributions. (b) The data-driven class distribution determined by examples falling within it. (c) A NDT without class hierarchy, where the leaves are experts that perform arbitrary predictions.}
\label{fig:NDT_hierarchy}
\end{figure}

Because the limitations of post-hot analysis (such as approaches in \hyperref[sec:DT-interpreting-NN]{section 2.2}), people might prefer to find \textit{ante-hoc} models that are designed to be inherently and intrinsically interpretable. In the filed of DT-based interpretations, neural decision trees (NDTs) are such ante-hoc models that are receiving increasing interest in recent years. Unlike approaches introduced in the last section, NDTs implement both the class hierarchy and the decision branches, so that they are perceived as full integration of the two worlds. Their core idea is to exploit NNs in the tree design by making the routing functions differentiable, thus allowing gradient descent-based methods to optimize. We first categorize NDTs according to whether it implements a class hierarchy (\textit{c.f.,} \autoref{fig:NDT_hierarchy}). If a tree satisfies this property, it means each internal node is assigned a specific and intermediate classification task. This “divide-and-conquer” strategy and stepwise inference process make the model more interpretable, because each node in the tree is responsible and distinguishable from other nodes. On the contrary, NDTs without class hierarchies restrain themselves little and perform arbitrary predictions at the leaves. Their the leaf nodes are usually classifiers rather than determined classes or distributions, so we dub them as \textit{expert NDTs} with expert leaves.

For approaches that implement a class hierarchy, we further classify them by if the architecture is data-driven. Data-driven methods employ data-dependent heuristics to perform local optimization. The resultant tree will lead to a recursive partitioning of the input space $X$ through a cascade of tests \cite{kuncheva2014combining, balestriero2017neural} and define the output of each leaf in terms of examples falling within it. We termed them as \textit{data-driven NDTs}. On the other hand, those without data-driven architectures tend to have a pre-defined structure and determine the leaf class distributions by priors or algorithms such that induce the class hierarchy. We call them \textit{bigot NDTs} with bigot leaves.

\subsection{Compositions and Inference Schemas of Neural Decision Trees \label{sec:inference-schema}}
\subsubsection{\textbf{Compositions}}
\begin{figure}
\centering
\includegraphics[width=\textwidth]{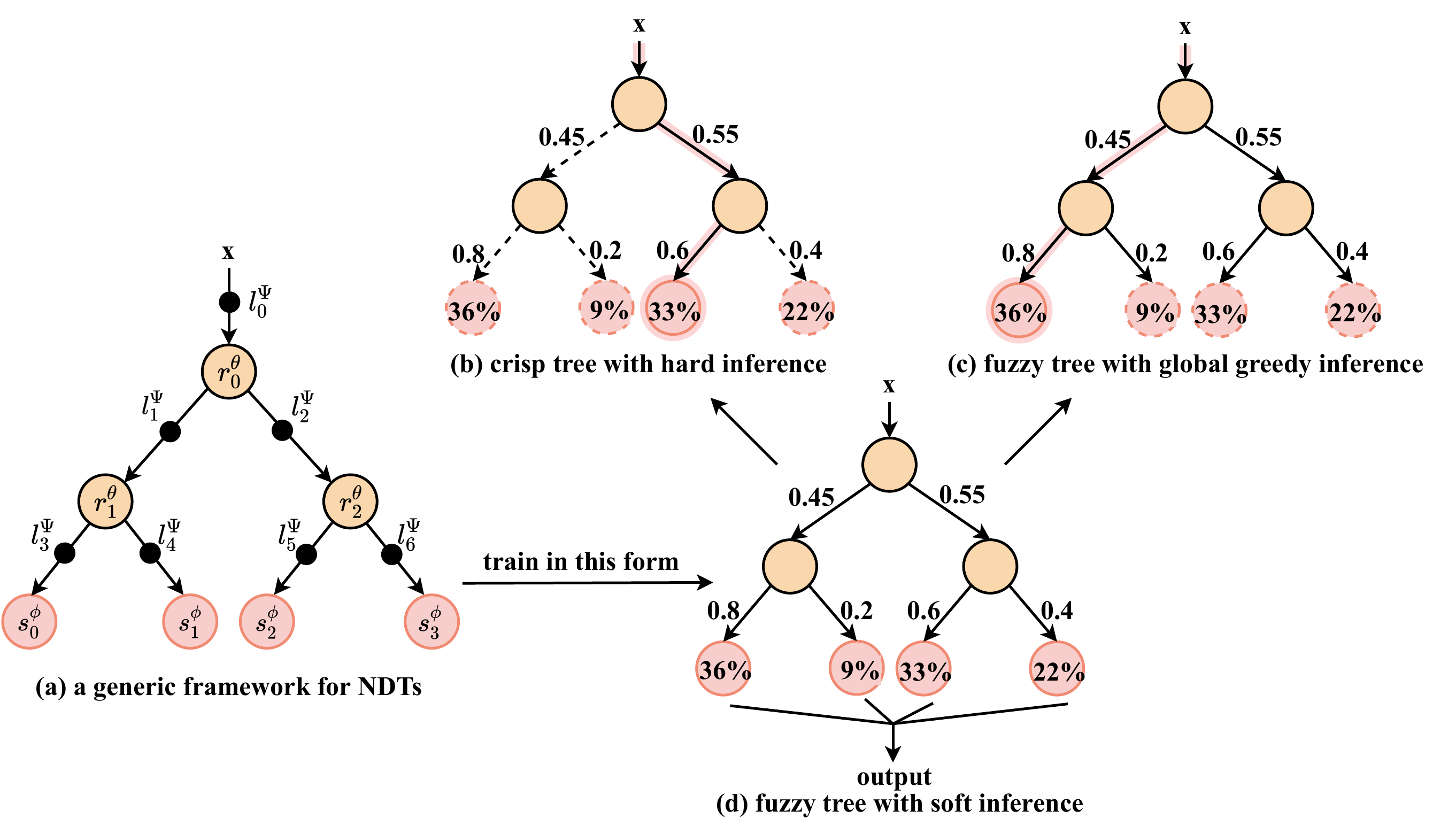}\\
\caption{A generic framework and different inference schemas for NDTs. (a) A generic NDT whose components are similar to \cite{tanno2019adaptive, chen2021self} (\textit{i.e.,} learners $l^{\psi}_{i}$, routers $r^{\theta}_{i}$ and solvers $s^{\phi}_{i}$), but the leaf solvers can be bigots. (b), (c) and (d) are three different inference schemas discussed in \hyperref[sec:three-inference-schemas]{section 4.1.2}}
\label{fig:generic-NDT}
\end{figure}

Considering a classification problem with training set $\mathcal{T}$ composed of $N$ labeled samples $(\mathbf{x}^{(1)}, y^{(1)}), \dots, (\mathbf{x}^{(N)}, y^{(N)}) \in \mathcal{X} \times \mathcal{Y}$, an NDT aims to obtain the prediction $\mathbf{\hat{y}}$ (defined as class probability distribution) for an input example $\mathbf{x}$. As illustrated in \autoref{fig:generic-NDT}, the model topology of a generic NDT can be defined as $\mathbb{T} := \left \{ \mathcal{N}, \mathcal{E}  \right \}$, where $\mathcal{N}$ is the set of nodes
and $\mathcal{E}$ is the set of edges between them. Each internal node $i$ has several children nodes, represented by $C(i)$. There are three primitive operations defined on $\mathbb{T}$ and each is conducted by a certain module.

\noindent $\bullet$ \textbf{Learners.} Every edge of the tree $i \in \mathcal{E}$ is assigned with a learner $l^{\psi}_{i}$, parameterized by $\psi$. A learner is tasked to transform samples from the parent and pass them to the children. However, the learner is not a necessary module in the NDTs. Only a few approaches adopt the nonlinear function as their learner for representation learning \cite{tanno2019adaptive, chen2021self}, yet most NDTs just use the identity function where the features stay unchanged when passed to later modules.

\noindent $\bullet$ \textbf{Routers.} Every internal node $i \in \mathcal{N}$ of the tree is assigned with a router $l^{\psi}_{i}$, which performs the routing function to send data to the children, and thus partitioning the input space, \textit{i.e.,} $r^{\theta }_{i} := \mathcal{X}_{i} \to [0, 1]^{N_{i}}$, parameterized by $\theta$. Here $\mathcal{X}_{i}$ denotes the input data space for router $r^{\theta }_{i}$, and ${N_{i}}$ is the number of children this router have. The router is a compulsory module in NDTs, and we can directly relate the way it processes with the reasoning mechanism of the model.

\noindent $\bullet$ \textbf{Solvers.} Solvers are leaf predictors that are supposed to give the final result, \textit{i.e.,} $s^{\phi}_{i} := \mathcal{X}_{i} \to \mathcal{Y}$, parameterized by $\phi$, where $\mathcal{Y}$ denotes the output space. In this survey, the solver is considered an essential module in NDTs. Its design (expert or bigot) greatly influences the performance and interpretability of the model and directly determines if the model implements a class hierarchy. If the conditional distribution $p(\mathbf{\hat{y}}|\mathbf{x})$ is varied in terms of the input example $\mathbf{x}$ (\textit{i.e.,} solvers are expert classifiers), the leaves will lose the ability of representing concrete concepts. Instead, if $p(\mathbf{\hat{y}}|\mathbf{x})$ is stationary (\textit{i.e.,} solvers are bigots, and the class distribution is determined by or identical to $\phi$), the leaves will be assigned with concrete concepts and thus inducing the class hierarchy.

\subsubsection{\textbf{Inference Schemas} \label{sec:three-inference-schemas}}
After installing the above modules, we consider three generic inference schemas that can be applied in a NDT.

\noindent $\bullet$ \textbf{Fuzzy tree with soft inference.} In a fuzzy tree, the input $\mathbf{x}$ stochastically traverses the NDT based on the probabilistic decisions (fuzzy memberships) provided by the routers until it reaches the leaves. In this case, the predictive distribution $\mathbf{\hat{y}}$ is calculated by averaging the prediction over all the solvers, weighted by their respective probabilities of reaching the leaves \cite{frosst2017distilling}. Supposing that there are $L$ leaf nodes, use $\Theta := (\theta, \psi, \phi )$ and $R$ to denote the involved parameters and the index of the reached leaf, respectively, then the full predictive distribution is determined by:
\begin{equation}
    p(\hat{\mathbf{y}}|\mathbf{x}, \Theta) = \sum_{l=1}^{L} p( R = l | \mathbf{x}, \psi, \theta  ) p(\hat{\mathbf{y}}|\mathbf{x}, R=l, \phi, \psi ),
    \label{equation6}
\end{equation}
where the first term $p( R = l | \mathbf{x}, \psi, \theta  )$ indicates the probability of reaching the $l$-th leaf, and the second term $p(\hat{\mathbf{y}}|\mathbf{x}, R=l, \phi, \psi )$ represents the predictive distribution produced by the $l$-th solver \cite{tanno2019adaptive}. For simplicity, we use $\mathcal{P}_{l}$ to denote the sequence of nodes along the unique path from the root to the $l$-th leaf, and $\pi^{\psi, \theta}_{l}(\mathbf{x})$ to denote the path probability (\textit{i.e.,} the first term of \autoref{equation6}) that is calculated by the product of decision probabilities over all router modules in path $\mathcal{P}_{l}$,
\begin{equation}
    \pi^{\psi, \theta}_{l}(\mathbf{x}) = \prod_{r^{\theta }_{i} \in \mathcal{P}_{l}}^{} \sum_{j=1}^{N_{i}} (r^{\theta }_{i}(\mathbf{x}^{\psi}_{i}))_{j} \cdot \mathbb{I}(C(i)_{j} \in \mathcal{P}_{l}),
    \label{equation7}
\end{equation}
where $\mathbf{x}^{\psi}_{i}$ is the representation of example $\mathbf{x}$ at node $i$. If the learner is an identity function, $\mathbf{x}^{\psi}_{i}$ will be the example itself. The indicator function $\mathbb{I}(C(i)_{j} \in \mathcal{P}_{l})$ output 1 only if leaf $l$ is in the $j$-th subtree of internal node $i$, otherwise output 0. Similarly, we can rewrite the second term of \autoref{equation6} into:
\begin{equation}
       p(\hat{\mathbf{y}}|\mathbf{x}, R=l, \phi, \psi ) = s^{\phi }_{l}(\mathbf{x}^{\psi}_{l}).
\end{equation}
If $s^{\phi }_{l}$ is an expert classifier, the output $\mathbf{\hat{y}}$ will be an arbitrary class distribution, otherwise it will output a static distribution which is determined by $\phi$ or is $\phi$ itself.

\noindent $\bullet$ \textbf{Fuzzy tree with global greedy inference.}
Instead of allowing each leaf to make a partial contribution to the final prediction, this inference schema uses the class distribution from the leaf with the greatest path probability, \textit{i.e.,} rewriting \autoref{equation6} into:
\begin{equation}
    p(\mathbf{\hat{y}}|\mathbf{x}, \Theta) = p(\mathbf{\hat{y}}|\mathbf{x}, R=\mathop{\arg\max}\limits_{l \in L} p( R = l | \mathbf{x}, \psi, \theta  ) , \phi, \psi ).
\end{equation}
It means calculating path probability for each leaf in terms of \autoref{equation7} and picking the solver with the highest one to give the prediction.

\noindent $\bullet$ \textbf{Crisp tree with hard inference.}
The above inferences adopt fuzzy splits with fuzzy membership functions to allow the example to traverse each child with respect to its probability. Crisp trees instead apply hard, deterministic inference to greedily traverse the tree, \textit{i.e.,} choosing the child with the highest probability at each internal node. This "single-path inference" strategy enjoys the lowest computation cost and tends to be more interpretable. However, it will result in some compromise to the accuracy. As shown in \autoref{fig:generic-NDT}, hard inference is prone to local optima.

\subsubsection{\textbf{Generic Training Procedure}}
For a NDT that is not data-driven, it typically has a pre-determined architecture. In order to permit the use of back-propagation and conduct global optimization, these models are usually designed as fuzzy trees with soft inferences during training phase. Once the training is over, the soft inference can be converted to more interpretable schemas with some sacrifice to accuracy. This is because soft inference needs to interpret the whole tree, involving routers at all inner nodes, while other schemas only need to interpret a list of routers along one particular path. On the other hand, data-driven NDTs need to grow from scratch. It stepwise trains each neural-based router, and assigns examples to its children with either crisp split or fuzzy split. Finally, it labels leaves with their respective sample distributions.

\subsection{Neural Decision Trees with Class Hierarchies}
In these approaches, class hierarchies will be determined in different ways, \textit{e.g.,} data-driven architectures, learned distributions or pre-defined pure classes. Similar to typical DTs, each internal node indicates a set of labels and may represent a superclass without a semantic meaning.

\subsubsection{\textbf{Data-Driven NDTs}\label{sec:NDT_with_data_driven}}
Data-driven NDTs are close to traditional DTs in the tree-inducing process. That is, if the node splitting takes place, a routing function defined on its input space is determined to satisfy a certain criterion, evaluated in terms of the examples passed to that node. In typical DTs, this is carried out in an approximate way. They randomly create a pool of parameter-free routing functions, afterwards retain the best one according to some class purity measure. Some data-driven NDTs likewise implement informativeness-based splitting functions with NN-based routers, which can be traced
back to three decades \cite{stromberg1991neural, guo1992classification, zhao2001evolutionary, zhou2002hybrid}. Stromberg \textit{et al.} \cite{stromberg1991neural} suggest training a small NN at each internal node. Subsequently, a natural choice from an indicator function will split the training data in order to single out one unique class that gains the most class purity. Similar strategies are proposed by Zhao \textit{et al.} \cite{zhao2001evolutionary}, where each node is a small NN that can be designed using genetic algorithms to maximize the information gain ratio. Particularly, zhou \textit{et al}. \cite{zhou2002hybrid} find that DTs prefer instances with unordered attributes, while NNs are vice versa. So they divide the instance space into ordered attributes and unordered attributes. They first induce a C4.5-like tree with unordered attributes. If a node can not further distinguish examples, it will be learned by a small NN trained with ordered attributes. Furthermore, Guo \textit{et al}. \cite{guo1992classification} partition the classes into two clusters at each internal node by two stages: find a pair of aggregate classes that minimizes a Gini impurity criterion, then find a good split between the two aggregations (through back-propagation), hence obtaining a good overall split.

Other approaches, however, do not improve the class purity when splitting the training data. They instead adopt other criteria. Sankar \textit{et al}. \cite{sankar1990fast, sakar1993growing, sankar1991optimal, sankar1991speaker} aim to decrease the classification error at each node to be extended. Besides, they extend the optimal pruning algorithm of binary classification trees \cite{breiman2017classification} to their multi-way tree design so as to improve its generalization. \cite{behnke1998competitive} introduces a different search method for data-driven NDTs, named the competitive neural trees (CNeTs). CNeT grows during competitive learning by using prototypes that represent clusters of examples at each node. Bengio \textit{et al}. \cite{bengio2010label} propose to map examples into the label embedding space and then predicts using a NDT. Unlike techniques we mentioned in \hyperref[sec:label-embedding]{section 3.3}, their approach does not impose the class hierarchy on the label embeddings. Instead, they assign each node of the NDT with a set of labels to which an example should belong if it arrives at that node. Deng \textit{et al}. \cite{deng2011fast} later develop this work to simultaneously determine the structure of the NDT and learn the classifiers for each node.

A few work resort to applying data-driven architectures to fuzzy NDTs. Suarez \textit{et al}. \cite{suarez1999globally} superimpose a fuzzy structure over the skeleton of a CART tree and introduce a global optimization algorithm that fixes the parameters of fuzzy splits. \cite{irsoy2014budding} likewise proposes a FDT model that can be dynamically adjusted. Each node starts as a leaf node, can then grow children, but later on these children can be pruned if necessary. \cite{laptev2014convolutional} applies fuzzy NDTs to image segmentation by extracting the most informative and interpretable features as the convolution kernels. Each split in the tree is represented by such a convolution kernel, which is learned in a supervised manner and used to maximize the informativeness of the split. More recently, \cite{balestriero2017neural} allows the use of DNNs with a new \textit{deep hashing layer} that is related to the Locality Sensitive Hashing (LSH) \cite{gionis1999similarity, charikar2002similarity}, an approach for mapping similar inputs to the same hash value. While the above FDTs prefer to realize fuzzy splits by replacing the typical indicator routing function with a sigmoid function, \cite{bhatt2006neuro} instead employs the \textit{fuzzy ID3 algorithm} that utilizes the \textit{fuzzy classification entropy} of a possibilistic distribution for DT-renovation. Similar to \cite{suarez1999globally}, they first use a induction algorithm to generate the DT, then fuzzify and optimize it using back-propagation algorithm while keeping the architecture intact.

Several approaches also concern the incremental learning capability of the NDT. Zhou \textit{et al}. \cite{zhou2002hybrid} provide three kinds of incremental learning. Two of them are designed for example-incremental tasks, while another one is a hypothesis-driven constructive induction mechanism. Irsoy \textit{et al}. \cite{irsoy2012soft} design a sigmoid-based FDT, where each split is made by checking if there is an improvement over the validation set. Su \textit{et al}. \cite{su2007neural} similarly propose to choose a target class at each internal node and train a small NN to separate the positive patterns from negative ones. Training patterns near the resultant hyper-ellipse decision boundaries are collected for the incremental learning.

The primary purpose of data-driven NDTs is to improve the design of typical DTs by differentiating parameters, extracting nonlinear features and decreasing the tree size, especially for complex problems that require highly nonlinear decision boundaries. However, because they retain the data-driven architecture like typical DTs, they may share the same limitations: their leaf labels are defined on the distributions of samples falling within it, and directly rely on the results of the input space partition. This learning schema can not represent the underlying trends in the data well and will probably limit the generalization capability. The lack of global optimization and enough representation learning also leads to unsatisfactory performance.

\subsubsection{\textbf{Bigot NDTs}\label{sec:NDT_with_determined_leaves}}
Instead of compromising to use a small NN at each tree node, bigot NDTs use one large NN to capture all regions in the feature space. They use pre-defined architectures and carry out global optimization. Each leaf either picks one pure class or learns a distribution that is one-hot or near one-hot. Besides, leaves stay unchanged during inference, which implicitly induces the class hierarchy.

We first introduce bigot NDTs with pure leaves, where the leaf labels are often randomly initialized or determined by prior knowledge. Sethi \textit{et al}. \cite{sethi1997structure} operate on a fixed structure of complete binary tree. These trees are initialized with empty internal nodes without routing functions, and terminal nodes are first labeled alternately as class 1, class 2, class 3, $\dots$, leading to an equal number for each class. Afterward, a learning scheme combining back-propagation with competitive learning is used to determine suitable splits for internal nodes and control the number of winning terminal nodes. The resultant tree is effective and compact, but the leaves are initialized casually and may cause unreasonable high-regularized space partition. Instead, Wan \textit{et al}. recently \cite{wan2020nbdt} propose the neural-backed decision trees (NBDTs), which use WordNet \cite{miller1990introduction} to assign a specific concept to each terminal and intermediate node. NBDTs are thus endowed with the ability of zero-shot superclass generalization, and a similar example can be referred to in \hyperref[fig:evaluation]{Figure 8}.

In contrast to pure leaves, other approaches assign each leaf with a probability distribution defined over the output space. The learned distribution is one-hot or near one-hot, so that it can also represent a certain class. Frosst \textit{et al}. \cite{frosst2017distilling} use binary FDTs where each internal node has a learned filter and each leaf node has a learned static distribution over the possible output classes. Since it is necessary to use FDTs in the training phase, determining the leaf distributions becomes a global learning problem. In this work, they choose to jointly optimize leaves and other parameters via back-propagation. A similar strategy is adapted in \cite{leon2016policy}, where they use Monte Carlo approximation to expect the gradient of the objective function in which a loss term is responsible for allocating the classes in the leaves. However, researchers find that jointly optimizing leaves and other network parameters gives inferior classification results \cite{nauta2021neural}, which inspires some work to explore algorithms that update the leaves independently \cite{nauta2021neural, kontschieder2015deep, rota2014neural}. Kontschieder \textit{et al}. \cite{kontschieder2015deep} first note that solely optimizing the leaf parameters is a convex optimization problem and proposed a derivative-free strategy. They alternatively update the leaf distributions and other network parameters, where the former perform as hyperparameters that should be determined before training the latter. Similar to NBDTs, they feed the trees with latent representation learned by a trained deep CNN. Their model is termed as deep neural decision forests (DNDFs), where the final fully connected layer of the trained CNN is used to provide routing functions for all nodes in all trees of the forest. In their algorithm, the predictive class distribution of a leaf solver is directly its normalized parameter, and the optimal predictions for all leaves given the routing decisions are obtained by minimizing a convex objective. With the notations in \hyperref[sec:inference-schema]{section 4.1}, we have: $s^{\phi}_{l}(\mathbf{x}^{\psi}_{l}) = \sigma(\phi_{l})$, where $\sigma(\cdot)$ denotes the softmax function, and there will be the following update scheme for $\phi_{l}$ for all $l \in L$:
\begin{equation}
    \phi^{t+1}_{l} = \sum_{\mathbf{x},y \in \mathcal{T}  }^{} \left (\sigma (\phi^{t}_{l})\odot \mathbf{y} \odot \pi^{\psi, \theta}_{l}(\mathbf{x}) \right )\oslash \hat{\mathbf{y}},
    \label{equation10}
\end{equation}
where $\odot$ denotes element-wise multiplication, $\oslash$ is element-wise division, $\mathbf{y}$ is the one-hot representation of the class label $y$ and $\hat{\mathbf{y}}$ is the predictive class distribution. Similar leaf updating algorithms are also applied to regression problems in \cite{shen2018deep} and semantic image labeling in \cite{rota2014neural}. However, this learning scheme may be computationally expensive because the training data has to be looped through twice at each epoch. The first time is to update $\phi$ in terms of \autoref{equation10}, and the second time is to update $\theta$ and $\psi$ via back-propagation while keeping $\phi$ fixed. Recently, Nauta \textit{et al}. \cite{nauta2021neural} propose to do this more efficiently. After updating $\theta$ and $\psi$ for each mini-batch, they immediately compute $\hat{\mathbf{y}}$ and $\pi^{\psi, \theta}_{l}(\mathbf{x})$ for updating $\phi$, so that the training data are traversed only once for each epoch. It is worthy mentioning that their work relates closely to the \textit{prototype-based} ante-hoc methods \cite{chen2019looks} and therefore termed as \textit{ProtoTree}. \textit{Prototypes} are interpretable representations of prototypical parts for classes, from which the evidence can be combined to make the prediction. The ProtoTree incorporates a prototype into each node of the tree. This prototype will be trained and projected to its nearest latent patch present in the training data, so that the prototype can be visualized as an actual image patch. The presence or absence of this learned prototype in an image determines the routing decision made by a node, through which they can locally explain a single prediction by outlining a decision path of the tree.

Recently, several practices have been made to apply bigot NDTs to knowledge distillation. Song \textit{et al.} \cite{song2021tree} utilize the NDT to layer-wise dissect the decision process. They project features from different layers of a learned deep model (perform as the teacher) to a lower-dimensional subspace and compute average feature vectors to represent the corresponding categories. Afterward, the vectors are clustered and adopted to generate a NDT for interpreting the intermediate decision-making process of different layers. Finally, they endow the student model with the same problem-solving mechanism. Xue \textit{et al}. \cite{xue2021tree} likewise mine the underlying category relationships from a trained teacher network in a layer-wise manner. It determines the appropriate layer in the network on which specialized branches grow to reconcile the conflicting decision patterns of different classes. This derived branching network, namely the NDT student, will be learned by knowledge distillation. Frosst \textit{et al.} \cite{frosst2017distilling} propose to use a trained NN to provide soft targets for training a fuzzy NDT, where the leaves are class distributions learned by back-propagation. Similarly, Li \textit{et al}. \cite{li2020tnt} provide a distillable tree model with learned distributions at the leaves. In their workflow, traditional tree-based models are used to provide an embedding dataset on which a deep model will be trained later. Once the deep model is trained, it generates new soft labels to train a distillable tree model. In such a learning schema, the knowledge is alternately transferred between the tree model and DNNs. Notably, their proposed distillable tree model is \textit{distillable Gradient Boosted Decision Tree} (dGBDT). We do not introduce GBDT \cite{friedman2001greedy} in our survey because it performs gradient descent in function space rather than in parameter space, so it does not relate much to NNs. However, the dGBDT turns itself into an ensemble of NDTs, thus enjoying continuous, differentiable and learnable parameters.

\subsection{Neural Decision Trees without Class Hierarchies (Expert NDTs)}
Except for the leaves, there are few differences between expert NDTs and bigot NDTs. However, such a slight change will significantly influence the performance and interpretability of the tree models. Expert NDTs impose little regularization on the network architecture and perform arbitrary predictions at leaf solvers, which leads to the lack of class hierarchy and the ambiguous task for each node. The very first proposed approach within this category is the Hierarchical Mixtures of Experts (HME) \cite{jordan1994hierarchical}, a tree-structured model for regression and classification. In our view, the HME is a fuzzy NDT that employs probabilistic methods in both the way it splits the input space and the way it combines the outputs from the experts \cite{waterhouse1994classification}. It adopts linear classifiers as its routers and fixes its architecture before training commences. In its original formulation, the parameters are determined by the maximum likelihood that can be solved with gradient ascent, and the model is trained using the Expectation Maximisation (EM) algorithm in which the "missing data" is specified.

HME has inspired many approaches based on fuzzy probabilistic splits. In \cite{waterhouse1994classification}, each expert is non-linear and performs multi-way classification. \cite{bishop2012bayesian, waterhouse1995bayesian, ueda2002bayesian} attempt to provide Bayesian treatments of the HME model to prevent the severe overfitting that is caused by parameters determined by maximum likelihood. Besides, we found that many tree models are constructed in a HME form, namely the fuzzy splits in the input space and the combinations of outputs from the expert leaves. These approaches improve the original HME framework by adjusting the optimizing methods and the network components. More recently, scholars prefer to combine HME models with CNNs or other advanced frameworks \cite{ji2020attention, xiao2017ndt, ioannou2016decision, yang2018deep, kim2022vit}. Ji \textit{et al.} \cite{ji2020attention} propose to incorporate convolutional operations along edges of the tree, which learns to capture the representations of objects. They also use the \textit{attention transformer module} to enforce the network to capture discriminative features. Xiao \textit{et al}. \cite{xiao2017ndt} proposes to transform the input by the \textit{feature network}, and then the hidden features are classified by the NDT. Their work is special in that they reformulate the non-differentiable information gain in the form of \textit{Dirac symbol}, and approximate the Dirac symbol as a continuous function. Ioannou \textit{et al}. \cite{ioannou2016decision} propose a hybrid model between decision forests and convolutional networks, named \textit{conditional network}. It fuses efficient data routing with accurate data transformation in a single model, and the data routing can be implicitly implemented by using groups of feature maps so as to yield higher compute efficiency. Particularly, Yang \textit{et al}. \cite{yang2018deep} proposes to implement a binning function that takes as input a continuous variable $x$ and produces an index of the bins to which $x$ belongs. After binning each feature of the input instance, they are able to determine the leaf node where the instance will arrive and a linear classifier will be used to classify it. More recently, \cite{kim2022vit} proposes a transformer version of the ProtoTree, where a vision transformer (ViT) acts as the backbone that produces learned representation, followed by a NDT decoder that is designed to resemble ProtoTree but with expert leaves.

All the NDTs introduced before use identity transformers as learners along edges, and will be fed with learned representations when handling high-dimensional real-world problems. However, the simplicity of identity transformers means that the input data is rarely learned. On the contrary, there are a few methods \cite{tanno2019adaptive, chen2021self, ioannou2016decision} performing representation learning along each edge, and we find some of them are special in searching network architectures. For example, \cite{tanno2019adaptive} grows the NDT architecture by greedily choosing the best option between going deeper and splitting the input space. But similar to traditional DTs, it relies on a suboptimal progressive scheme to grow trees where each operation is selected in a greedy manner, making it easily prone to local optima. Instead, \cite{chen2021self} proposes NDTs that are self-born from a large search space, but still have some shortcomings. \textit{I.e.,} the number of parameters during the search phase is too large. Besides, the born NDT operates on a different schema from the search phase, because it removes some neural components and connections in the network and trains it from scratch. Moreover, the resultant architectures often degenerate into a single branch (\textit{i.e.,} a flat classifier). Our experiment of expert NDTs in \hyperref[sec:ndt-trade-off]{section 5.2} shows the same result and will be discussed later.

\subsection{Summary and Discussion}
NDTs are hybrid models of NNs and DTs, and are inherently interpretable. We demonstrated that the leaf design determines whether the model implements the class hierarchy and whether the model is interpretable in terms of \textit{decomposability}. For data-driven and bigot NDTs, each internal node is dedicated to classifying a concept at a certain level. However, their structures are thus highly regularized and the solution of a soft inference is limited within the linear combination of the leaf distributions, which greatly reduces their performance. On the contrary, expert NDTs are usually characterised by using expert leaves to boost performance, but suffer from the lack of class hierarchy. The only exception to our knowledge is \cite{ahmed2016network}, an approach that uses expert leaves while implementing the class hierarchy. Their idea is simple: the trunk of the network are convolutional layers optimized over all classes, but at a given depth, it splits into separate branches and each dedicated to discriminating a subset of classes. Their design is perceived as seeking an ideal trade-off between performance and interpretability, which will be discussed detailedly in the next section.

\section{Analysis and Prospects\label{sec:Analysis_Comparison}}
In the previous sections, we introduced three lines of NTs, and each ends with a summary of their contributions and limitations. These techniques are of extraordinarily different settings, developed in different times and applied to different fields. Despite the huge conflicts, we find most NTs can be treated as efforts of improving the model interpretability. Concretely, there are two main techniques: \textbf{NN-approximation methods} in \hyperref[sec:DT-interpreting-NN]{section 2.2} and \textbf{neural decision trees} in \autoref{sec:NDT}, which makes up the principal part of the taxonomy. Herein we are going to provide a general comparison for them, and offer more specific analyses for the fine-grained techniques and other possible interpretability-related approaches. Before the analysis, it is beneficial to indicate how to evaluate the interpretability of a model. We simply propose the "What" and "Why" criteria, implications of interpretability at two levels:

\noindent $\bullet$ \textbf{What: the explicit content of a model or a part of the model learns}. It shows the \textit{decomposability} of the model, which aims to understand a model in terms of its components such as neurons, layers, blocks and so on \cite{fan2021interpretability}. In a typical NN, "What" is a simple input-output mapping. There is only one functionalized module, \textit{i.e.,} the whole NN, and the "What" of the inner structure or an individual neuron is not accessible. On the contrary, DTs are a kind of well-modularized method, where each module has an explicit utility to perform primitive operations.

\noindent $\bullet$ \textbf{Why: the transparent process of learning a specific "What"}. It shows the algorithmic transparency for understanding the training and reasoning process of a model. For instance, in a node of a axis-aligned DT, "What" means it learns how to use a single feature to make the intermediate decision, while "Why" tells us how this "What" processes (\textit{e.g.,} by some informativeness-based criteria or other heuristics). For a typical NN, however, "Why" is entirely not available because it is encoded in the network parameters in a form not amenable to human comprehension.

There have been great efforts for exploring model interpretability and explainable AI, and we follow the idea of \cite{du2019techniques} and \cite{fan2021interpretability}, which treat post-hoc and ante-hoc interpretability separately. We indicate that NN-approximation methods in fact perform post-hoc analysis which explains the existing black-box models, while NDTs are ante-hoc modeling approaches that construct interpretable ones. This section will analyze the two techniques separately, and each ends with a discussion about the remaining challenges and future directions. Afterward, we will discuss DT-inspired NNs and indicate their relations to NDTs. For the rest approaches that account for a few part of this survey, we will offer a review of their "What" and "Why" in addition to the available summaries in previous sections. Finally, other considerations like conditional computation are discussed. It is noted that our analysis will focus on NDTs, because other approaches are proven to be implicit or half integration that are less interpretable or seldom developed in recent years. 

\subsection{NN-approximation Methods: Approximate "What" and Surmise "Why"}
\subsubsection{\textbf{Review of Three Techniques}}
Methods introduced in \hyperref[sec:DT-interpreting-NN]{section 2.2} interpret a trained NN by inducing a conventional DT to extract rules from it. Such a post-hoc strategy in fact performs "What" approximation (decompose and approximate decision boundaries learned by the NN) and "Why" surmise (the cause of forming such decision boundaries is surmised by the comprehensible and symbolic rules), but it is not capable of finding out the exact "What" and "Why" of the black box itself. These methods are varied in terms of how they characterize the internal structure of the network. The pedagogical techniques still treat NNs as black boxes and only need to extract rules for the simplest input-output mapping. They enjoy the most compact and effective trees at the expense of relatively low fidelity and performance. On the contrary, decompositional techniques are exhausted to extract rules for every neural link to preserve the inner structure well and achieve the highest fidelity. Eclectic techniques are the trade-off between them, which require access to partial knowledge in the black box but do not demand to simulate the "Why" of each neuron. More discussion can be referred to in the summary of \hyperref[sec:summary-approximation]{section 2.2.5}.

\subsubsection{\textbf{Other NN-approximation Methods}}
In addition to the techniques in the taxonomy, there are methods of approximating NNs that adopt other strategies or employ other machine learning models. They are mainly characterised by constructing proxies (surrogate models) to closely resembles black-box deep models \cite{fan2021interpretability, zhang2018visual}. We take pedagogical techniques for instance, which comprise the VIA methods \cite{thrun1994extracting, thrun1993extracting}, other sampling-based approaches \cite{sethi2012kdruleex, taha1999symbolic, zhou2000statistics} and more recently, the LIME techniques \cite{ribeiro2016should, garreau2020explaining}. VIA uses validity interval analysis to extract $if-then$ rules that can be expressed by arbitrary linear constraints. Other sampling-based approaches share the same idea of creating extra artificial training examples, but the algorithms for learning rules are carried out by other exemplary models like decision tables \cite{sethi2012kdruleex}, rule systems \cite{taha1999symbolic} instead of DTs. LIME particularly proposes a generic framework for approximating the predictions of any classifier or regressor with a simple interpretable model. However, this is an explaining-by-case method that only behaves in the vicinity of a given instance, and does not imply global fidelity. Compared with these approaches, the use of DTs is superior in that they are more intuitive and convenient for humans to understand the abstract rule knowledge implied in a large assemblages of real-valued network parameters. Similar comparisons can also be found in decompositional and eclectic techniques, where DTs are adopted much less than that in pedagogical techniques.

\subsubsection{\textbf{Discussions and Prospects}}
Most of these approaches are model-agnostic and flexible to explain future models, because they do not modify the original operation of the black-box model. However, in addition to the extra cost for constructing a proxy, these techniques also encounter a dilemma that it is often impossible to identify surrogate models that are globally and entirely faithful to the original one. Approaches like LIME compromise to protect the local fidelity by producing interpretations for each specific instance, but usually do not imply global fidelity. We ascribe this challenge to the simplicity of the surrogate models. An interpretable model tends to be simple and may not be capable of fitting highly non-linear decision boundaries. Therefore, we suggest employing surrogate models that are more complex and powerful (\textit{e.g.,} NDTs) without losing much interpretability. However, a worse limitation of these approaches is that we often do not know the nuance \cite{fan2021interpretability}, which makes it hard to have a full trust to a post-hoc interpretation because the correctness is not guaranteed. In this case, we may have to agree with \cite{rudin2019stop} in that stop explaining black-box models for high-stakes decisions and use interpretable models instead.

\subsection{Neural Decision Trees: Trade-offs Between Performance and Interpretability\label{sec:ndt-trade-off}}
NDTs in \autoref{sec:NDT} aim to address the black-box nature of deep models, instead of approximating such models. In addition to the different frameworks they operate on (\textit{e.g.,} MLP, CNN, Transformer \cite{vaswani2017attention, dosovitskiy2020image}), we concern more about the different designments of components in the tree. The essence of these differences is the trade-off between performance and interpretability, because different designments are endowed with different "What" and "Why". Before our analysis, we refer readers to \hyperref[fig:evaluation]{Figure 7} to get a general comprehension, where we design several NDTs for quantitative evaluation on the Cifar10 dataset. Similar to \cite{chen2021self, tanno2019adaptive}, we adopt small CNNs as learners and routers, and their backbones are two convolution layers followed by other regular operations like pooling or normalization. Leaf solvers are varied, either experts with linear classifiers or bigots with determined class distributions. The workflow of "Bigot1" is to design a large binary NDT and train it with soft inference, afterwards prune the branches that are seldom visited by examples such that reduce a lot of parameters. Based on the architecture generated by "Bigot1", we design other NDTs for controlled experiments. Details of these experiments will be described below.

\subsubsection{\textbf{Routers: Direct Implementer of "What" and "Why"}}
Every internal node of a NDT is assigned with a router that implements the routing function, from which the final prediction is obtained by a sequence of decisions. As for bigot NDTs and data-driven NDTs, each router implements a straightforward "What" that it learns to distinguish examples into subclasses by judging if a discriminative condition is satisfied or not. On the contrary, expert NDTs can not tell us what they exactly learn at each internal node. Despite the success of implementing "What" in some NDTs, we unfortunately find that seldom have these approaches implemented an apparent "Why". They often construct neural-type connections or small NNs as the router, so each decision is made by a small black box. We suggest that future work refer to the practice of ProtoTree \cite{nauta2021neural} and seek means or tricks to endow the router with a more transparent reasoning mechanism.

\begin{figure}[tb]
	\centering
	\begin{minipage}[h]{0.5\textwidth}
		\centering
		\includegraphics[width=0.97\linewidth]{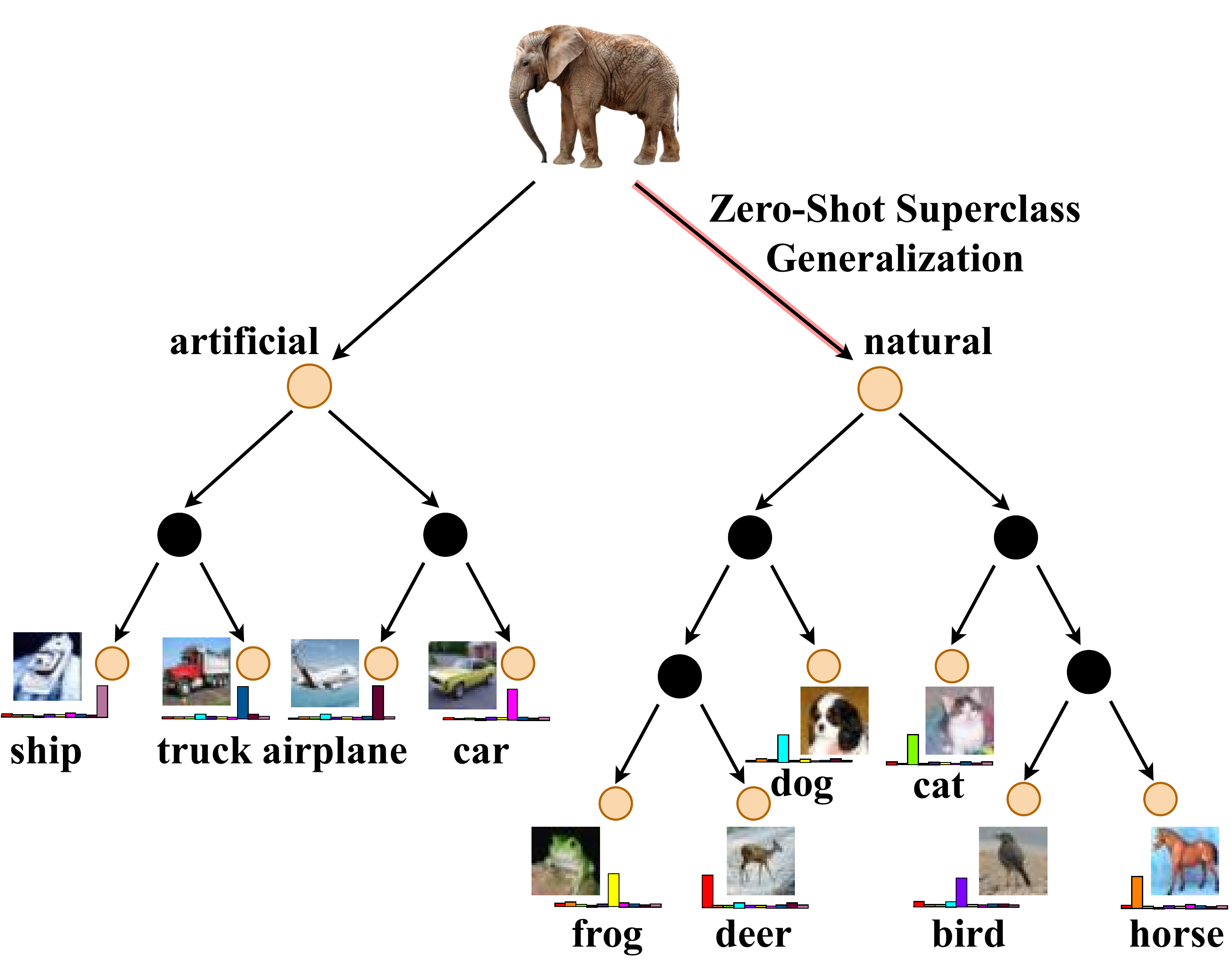}
		\label{fig:evaluation}
	\end{minipage}
	\begin{minipage}[h]{0.45\textwidth}
		\centering
		\includegraphics[width=0.97\linewidth]{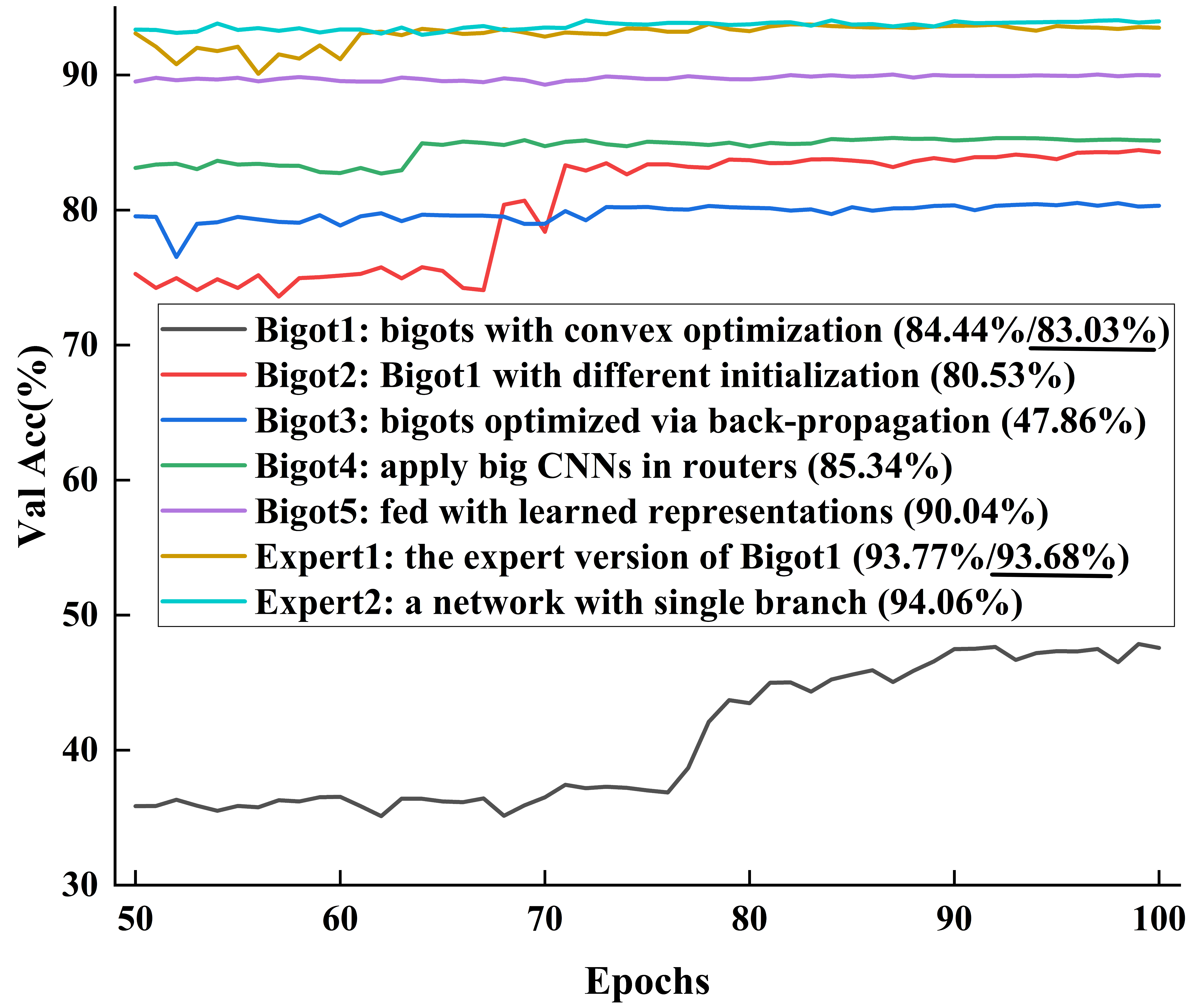}
		\label{fig:evaluation}
	\end{minipage}
\caption{(Left) A class hierarchy that is partially meaningful in taxonomy, corresponds to the "Bigot1" in the right figure. The light brown nodes can be assigned with a specific meaning and allow the superclass generalization, whereas dark nodes represent abstract concepts that can not be determined in terms of taxonomy or other known criteria. (Right) Quantitative evaluation for specially designed NDTs on the Cifar10 dataset. The underlined numbers denote the results of single-path hard inference.}
\end{figure}

\subsubsection{\textbf{Leaf Solvers: How to Teach The Routers? - Expert VS Bigot}}
For most NDTs whose architectures are pre-defined, the leaf solvers act as teachers for routers. Because the leaves need to be determined first so that the routers can learn how to partition the input space. During this process, people may prefer bigots because they can induce the class hierarchy and teach routers meaningful "What". Contrarily, experts will lead to uninterpretable decisions and stochastic routing paths. Despite the superiority of interpretability, bigots often result in a significant decline in accuracy and fail to jointly optimize with other parameters via back-propagation. These comparison can be referred to in "Bigot1", "Bigot3" and "Expert1" of \hyperref[fig:evaluation]{Figure 7}. The bigot leaves of Bigot1 are determined independently by the convex optimization algorithm in \cite{kontschieder2015deep, nauta2021neural}. Bigot3 instead jointly optimizes its leaves with other parameters via back-propagation. Expert1 is an expert NDT with the same architecture of Bigot1. We perform same initialization for the three NDTs (except the leaves in Expert1). The experiment shows that Expert1 greatly outperforms other bigot NDTs, and bigot leaves directly optimized by back-propagation will give the worst results.

We indicate that the poor performance of bigot leaves is caused by high regularization. In a bigot NDT, a class hierarchy is implemented where each intermediate decision has a far-reaching impact on the final prediction. Contrariwise, an expert NDT does not impose much regularization, and does not care about the correctness of the intermediate routing decisions as long as the expert leaves can predict right. With regard to the failure of back-propagation optimization in bigot NDTs, our conjecture is the limited participation of the bigot leaves during inference. As a part of the NDT, the leaf class distributions are independent from other network parameters. It is not produced by the learned representations and is not related to the previous neural links. Although the possible linear combinations of the static leaf distributions can cover the whole output space, including them in the loss term can lead to an overly complex optimization problem.

Several approaches propose convex optimization for updating leaves independently, but we find that the resultant model is unstable and affected by initialization. \textit{C.f.,} the comparison between "Bigot1" and "Bigot2" in \hyperref[fig:evaluation]{Figure 8}. Bigot1 \textbf{happens} to discover the class hierarchy that is partially meaningful in taxonomy. However, with another initialization (Bigot2), it induces a hierarchy without any taxonomically meaningful internal node and brings about a worse result. We suppose this learning schema leads to great constraints on the structure, forcing some classes to be similar, rather than they are inherently similar in taxonomic, semantics or other known criteria. Therefore, the induced class hierarchy may exacerbate the learning of the model. Even the derived class hierarchy is meaningful, it is believed that the human intuitions on relevant classes do not necessarily equate to optimal representations \cite{tanno2019adaptive}. We indicate that the definite advantage of a meaningful class hierarchy is the ability to perform superclass generalization. An example is Bigot1 (\textit{c.f.,} the left of \hyperref[fig:evaluation]{Figure 8}), where the elephant has not been presented to the NDT, but belongs to the superclass "natural" or "animal", so that a zero-shot generalization is available. Only superclasses with actual concepts can enjoy this generalization, because we lack access to evaluate abstract concepts.

It is also noted that these NDTs are all based on the architecture of Bigot1, derived from a large binary bigot NDT. If we turn the search space into a binary expert NDT, the resultant architectures will often degenerate into a flat classifier with a single branch (\textit{i.e.,} Expert2), which obtains the highest performance. This is the same result as our previous conclusion for \cite{chen2021self}, and we confirm it from its authors. We speculate this is due to the lack of class hierarchy and task-specific nodes. In this case, a flat classifier is already sufficient to accomplish the classification task. The employment of decision branches and probabilistic methods will instead complicate the optimization.

\subsubsection{\textbf{Formation of The Tree: Data-Driven VS Neural-Learning}}
Another aspect worth discussing is the formation process of a NDT, either data-driven or pre-determined as a NN. The two schemas resemble the confrontation of hierarchical cluster and hierarchical representation. Data-driven NDTs have fully available "What", and possible "Why" if the splitting of a decision node is intelligible (\textit{e.g.,} an indicator function in terms of the state of a feature). Such interpretability is absent in expert NDTs, but can be partially achieved (concerning "What") by bigot NDTs.

We suppose that using data clusters to determine the class distributions can not represent the underlying trends in the data well, and will probably limit the generalization. Besides, most data-driven NDTs are not continuous during the input-output inference, because they impose discrete constraints at the splits for hard partitions (\textit{i.e.,} trained as crisp trees). One possible solution is to perform global finetuning after the tree-inducing phase and fuzzify the routing functions to impose continuity constraints. \cite{suarez1999globally} is a practice of this strategy. But if we design the whole DT as a NN and globally optimize the pre-determined structure, these issues will be addressed more directly and effectively, meanwhile a higher performance will be achieved. For these reasons, data-driven NDTs have rarely developed in recent years.

\subsubsection{\textbf{Input of The Tree: Where to Start Interpretation}}
As we introduced in previous sections, a few approaches prefer to feed learned latent representation to the NDT in order to deal with high-dimensional real-world problems. Because there is less redundant information in the latent representation such as the correlation between pixels and channels, NDTs fed with latent representations will certainly perform better than those fed with raw data. The comparison between "Bigot1" and "Bigot5" in \hyperref[fig:evaluation]{Figure 8} can prove this. Note that we do not feed latent representation to "Bigot5", we instead input the feature maps obtained from an intermediate layer in the VGG-13 network. The result of "Bigot5" shows great advantages in performance over those fed with raw data. NDTs fed with latent representations will probably perform better and can even match its backbone network, \textit{e.g.,} NBDT based on ResNet18 can achieve 94.82$\%$ in Cifar10, close to its baseline 94.97$\%$. The result is not surprising, because the change of the deep model is slight, \textit{i.e.,} replacing the final linear layer with a NDT. In this case, the task for the NDT is to guide the representation learning in the lower layers of the deep model, whereas the representation learning aims to reduce the uncertainty of decisions taken at the routers. All the decisions produced by routers are in fact posterior probabilities after the parameters of the network backbone are determined. In our view, this greatly reduces interpretability because it fails to perform the full interpretation from the original input space to the final prediction. 

For NDTs fed with the raw data, however, it is often necessary to simultaneously perform stepwise dimension reduction (\textit{e.g.,} down-sampling for images) and decision-making in case the input is high-dimensional. Such a learning schema uses non-identity learners and is hardly scalable to a large dataset, because the number of parameters will roughly double each time the tree deepens. Besides, in most internal nodes, the representations have to make intermediate decisions before they are learned well, so it is apparent to sacrifice some performance. Although each router can be designed as an entire NN that is input with the raw data and output routing probabilities, this will result in a rather cumbersome backbone and can only be applied to small datasets. The only practice to our knowledge is \cite{frosst2017distilling}, a bigot NDT for handling the MNIST dataset. "Bigot4" is our design for this practice. We replace the small CNN in each router with a whole VGG-13 network, which slightly improves the performance at the expense of a tremendous increase in parameters.

\subsubsection{\textbf{Empirical Comparison}}

\begin{table}
\centering
\caption{Qualitative comparison between a part of representative work of different NDTs, with respect to their interpretability. S. path indicates the single path inference, where the work with marked ticks can satisfy this property during both the training and testing phase, and the marked ticks in "Bigot" imply data-driven architectures. "ens.", "Inter" and "Trans." are short for ensemble, interpretation and Transparency, respectively.}
\begin{tabular}{m{0.16\textwidth}<{\centering}m{0.07\textwidth}<{\centering}m{0.08\textwidth}<{\centering}m{0.08\textwidth}<{\centering}m{0.08\textwidth}<{\centering}m{0.07\textwidth}<{\centering}m{0.08\textwidth}<{\centering}m{0.08\textwidth}<{\centering}m{0.07\textwidth}<{\centering}}
\hline
{\textbf{\rotatebox{60}{Name}}} & {\textbf{\rotatebox{60}{Year}}} & 
{\textbf{\rotatebox{60}{S. path}}} & {\textbf{\rotatebox{60}{Bigot}}} & {\textbf{\rotatebox{60}{Non-ens.}}} & {\textbf{\rotatebox{60}{Full Inter.}}} &
{\textbf{\rotatebox{60}{Genuine $l^{\psi}$}}} &
{\textbf{\rotatebox{60}{Trans.}}} & {\textbf{\rotatebox{60}{Others}}}                               \\ 
\hline
NT \cite{stromberg1991neural} & 1991 & $\checkmark^{*}$ & $\checkmark^{*}$ & \checkmark & \checkmark & $\times$ & $\times$ & $\times$ \\
\rowcolor{mygray} HME \cite{jordan1994hierarchical} & 1994 & \checkmark & $\times$ & \checkmark & \checkmark & $\times$ & $\times$ & $\times$ \\
— \cite{suarez1999globally} & 1999 & \checkmark & $\checkmark^{*}$ & \checkmark & \checkmark & $\times$ & $\times$ & $\times$ \\
\rowcolor{mygray} QUANT \cite{su2007neural} & 2007 & $\checkmark^{*}$ & $\checkmark^{*}$ & \checkmark & \checkmark & $\times$ & $\times$ & $\times$ \\
DNDF \cite{kontschieder2015deep} & 2015 & $\times$ & \checkmark & $\times$ & $\times$ & $\times$ & $\times$ & $\times$ \\
\rowcolor{mygray} SDT \cite{frosst2017distilling} & 2017 & \checkmark & \checkmark & \checkmark & \checkmark & $\times$ & $\times$ & $\times$ \\
ANT \cite{tanno2019adaptive} & 2019 & \checkmark & $\times$ & \checkmark & \checkmark & \checkmark & $\times$ & \checkmark \\
\rowcolor{mygray} AC-Net \cite{ji2020attention} & 2020 & \checkmark & $\times$ & \checkmark & $\times$ & $\times$ & $\times$ & $\times$ \\
SeBoW \cite{chen2021self} & 2021 & \checkmark & $\times$ & \checkmark & \checkmark & \checkmark & $\times$ & \checkmark \\
\rowcolor{mygray} NBDT \cite{wan2020nbdt} & 2021 & \checkmark & \checkmark & \checkmark & $\times$ & $\times$ & $\times$ & \checkmark \\
ProtoTree \cite{nauta2021neural} & 2021 & \checkmark & \checkmark & \checkmark & $\times$ & $\times$ & \checkmark & \checkmark \\
\rowcolor{mygray} Vit-Net \cite{kim2022vit} & 2022 & \checkmark & $\times$ & \checkmark & $\times$ & $\times$ & \checkmark & \checkmark \\
\hline
\end{tabular}
\label{tab1}
\end{table}

After the quantitative evaluation for our specially designed NDTs, we give an empirical comparison between some representative work of different NDTs in \autoref{tab1}, with respect to the following qualitative indicators of interpretability: (1) \textbf{Single-path inference} means carrying out discrete and sequential decisions so as to select one path and pick one leaf solver. (2) \textbf{Bigot} with pure/determined leaves, so that one leaf picks a particular class or a determined class distribution. (3) \textbf{Non-ensemble}, so that path to prediction attribution is discrete \cite{wan2020nbdt}. (4) \textbf{Full interpretation}, so that NDTs are fed with the raw data and responsible for the full interpretation from the original input space to the final prediction without any prior knowledge. (5) \textbf{Genuine learner $l^{\psi}$}, where the transformation functions of learners are not the simple identity-mapping, so that the tree can stepwise learn representation and simultaneously make decisions. (6) \textbf{Transparent routing} focus on "Why", so that the routers do not simply use small NNs with a black-box nature to make decisions, they are instead endowed with a transparent reasoning process. (7) \textbf{Other possible contributions for interpretability}, such as reasonable architecture search \cite{chen2021self, tanno2019adaptive}, zero-shot superclass generalization \cite{wan2020nbdt}, reasoning with prototypes \cite{nauta2021neural} and so on. The ticks mean preserving these properties, while crosses desert them so as to obtain higher accuracy.

It is noted that the exact quantitative evaluation of performance between these approaches is impossible or makes little sense. Because most NDTs are specialized and not portable. For instance, in a data-driven NDT or a bigot NDT, the number of classes determine the minimum number of leaves, so NDTs designed on large dataset like ImageNet will grow to an unacceptable depth. Generally, only expert NDTs fed with learned representations are flexible enough, but their network backbone (\textit{e.g.,} MLP, CNN, Transformer) will also have a significant impact on their performance. With these concerns, we do not include the performance evaluation in \autoref{tab1} even though they are roughly consistent with our previous hypotheses.

\subsubsection{\textbf{Perspectives of NDTs}}
As discussed above, some NDTs realize interpretability at the expense of excessive constraints and sacrifice of performance. This section aims to find more ideal trade-offs or sidestep such trade-offs, in hope to advance the practice of NDTs. 

\noindent $\bullet$ \textbf{Use experts with class preference.} We found that few NDTs have been proposed to simultaneously implement the class hierarchy and deal with the high-dimensional raw data for full interpretation. The lack of such practice may be ascribed to the high regularization imposed by the bigot leaves, which force classes to be similar. We suggest a possibility of implicitly inducing class hierarchies by softening bigots into \textit{experts with class preference}. That is, we may perform a convex optimization on the weight matrix $\phi$ of the leaf solver (here is an expert with a linear classifier) while keeping other parameters fixed. We adjust the connections $\phi^{(i)}$ to have the sparsity property $\phi^{(i,j)} \approx 0$ or the negative relation $\phi^{(i,j)} < 0$ if the weights of the $i$-th row do not contribute to the class the expert prefers. This sparsity or negative contribution is desirable because it encourages experts to make predictions relying less on their disliked classes \cite{chen2019looks}. In this way, experts are allowed to represent a particular class, but do not impose a mandatory constraint on the network. Experts can also participate in the back-propagation to be adjusted after the convex optimization. The specific strategy needs determining by future practices. If the NDTs employ priors like WordNet to label the class hierarchy, the weights of experts may be initially fixed at a small negative number. However, WordNet focuses on conceptual rather than visual similarity and lack concepts that are not themselves objects \cite{wan2020nbdt}. Besides, the class hierarchy derived from human intuitions does not usually perform well, especially for datasets without an underlying class hierarchy (\textit{e.g.,} MNIST) \cite{tanno2019adaptive}. Therefore, we suggest a self-born hierarchy for NDTs: randomly initialize the parameters of the expert, and define its preference in terms of the class distribution of examples falling within it. Afterward, alternatively perform gradient descent and convex optimization, of which the latter aims to enhance the class with the highest preference by weight tuning. More sophisticated methods for forming or leveraging hierarchies are left to future work.

\noindent $\bullet$ \textbf{Adopt network deconvolution for stepwise reasoning.} Another possible help is specific for NDTs fed with high-dimensional raw images. The existing methods normally use small CNNs as learners and routers. However, convolutional kernels are in effect re-learning redundant data \cite{ye2019network} because of the strong correlations in real-world images. Ye \textit{et al.} \cite{ye2019network} propose the \textit{network deconvolution} to optimally remove pixel-wise and channel-wise correlations. Considering a standard convolution operation that can be formulated into one large matrix multiplication $Xw$, where $w$ is the flattened 2D kernel and each column of $X$ corresponds to one flattened patch of a feature. $X$ is then used to calculate the covariance matrix $Cov = (X^{T}X / N)$, where $N$ is the number of samples from a batch. Afterward, the inverse correction for eliminating correlations is applied by multiplying $X$ with $Cov^{-0.5}$ before the weight training of $w$ (\textit{i.e.,} $y = X \cdot Cov^{-0.5} \cdot w$). We suggest applying network deconvolution in NDTs which demand rudimentary features make intermediate decisions. This will probably boost the performance of NDTs with the help of more accurate routing. For the detailed implementation of the network deconvolution, we refer readers to \cite{ye2019network}.

\noindent $\bullet$ \textbf{Design routers with transparent reasoning process.} With regard to the routers, one direction of the future work is to develop ProtoTree \cite{nauta2021neural}, a bigot NDT that enjoys a transparent decision-making mechanism owing to the use of prototypes. But such decisions only occur in the latent space, and we lack means of interpreting how the raw data are mapped to the latent space. Therefore, we suggest the possibility of designing NDTs that uses prototypes at different grains to perform stepwise reasoning. That is, at each time we reduce the dimension, we use the prototypes in that space to interpret decisions made by the routers, rather than only interpret in the latent space. Besides, the use of prototypes in ProtoTree is kind of unreasonable in that they project the prototypes into one particular patch of an image in the training set. Although this trick leads to impressive visualization, it is irrational for a patch to represent all the distinguishing features of one class. More importantly, in our experiments we find some prototypes were projected to the image background rather than the part of interests, which makes the reasoning process unreasonable in forcing classes to be related to irrelevant information. Therefore, we suggest renovating prototypes by using more patches, sidestepping irrelevant backgrounds or changing the mapping approaches to represent more reasonable prototypical parts of a class. Absolutely, we do not have to use prototypes to endow NDTs with a transparent reasoning mechanism. Other efforts of incorporating ante-hoc interpretation for routers or other neural components are also worth practicing.

\noindent $\bullet$ \textbf{Draw lessons from other NN-renovation methods.} 
NDTs are a kind of \textit{model renovation} method in ante-hoc interpretability modeling, which aims to deploy more interpretable machineries into a black-box model. After incorporating the statements in \cite{fan2021interpretability} and complementing them with our own views, these machineries include: (1) \textit{purposely designed neurons} \cite{chu2018exact, fan2017revisit}, so that the decision boundaries of the neurons are explicitly defined, thereby deriving closed-form solutions for predictions. (2) \textit{layers with special functionalities} \cite{fan2020soft, kuo2019interpretable}, so that the network can be interpreted as a learned cascaded model consisting of transparent layers. (3) \textit{modularized architectures} \cite{li2018deep}, where the learning of each component is steered towards a specified task. (4) \textit{models with simplified or transparent functions} \cite{vaughan2018explainable, jiang2020cold}, which are dedicated to simplifying partial derivatives or transforming NNs into more interpretable rule-system forms. (5) other strategies such as using hybrid predictive system \cite{wang2019gaining}. Among these NN-renovation methods, NDTs belong to "modularized architectures". They modularize NNs at the level of each neural component so that a better "What" can be achieved. Another example within this survey is the method introduced in \hyperref[sec:DT-structural-prior]{section 2.1}, which uses DTs as structural priors to modularize NNs into specialized and crafted layers. This is advantageous to the optimization of the network design since we know the role of each component. We suggest employing other kinds of NN-renovation methods in the design of NDTs, \textit{e.g.,} using more simplified and transparent functions for routers or learners, assigning other special functionalities to the neural components, or designing other interpretable modules for diverse tasks.

\subsection{\textbf{DT-inspired NNs: The Half Integration and Generalized NDTs} \label{sec:half-integration}}
In \hyperref[sec:intro-class-hierarchy]{section 1.4.2} and \autoref{sec:NN_with_hierarchy}, we introduced NNs that draw on a part of ideas from DTs. Because the absence of an actual DT in these approaches, they are perceived as half and implicit integration of the two worlds. As for NNs leveraging class hierarchies, we indicated their motivation is to employ the class hierarchy to improve performance and mitigate the severity of mistakes. Particularly, those with hierarchical architectures have more than one functionalized module, thereby increasing the interpretability in terms of "What". We did not include NNs utilizing decision branches in our taxonomy, because their decision mechanisms are quite different from those in typical DTs. Therefore, these approaches do not relate directly to the DTs, but we indicate they are associated with expert NDTs where the implementation of a class hierarchy is not demanding. A branched network and an expert NDT share the same components of routers and expert leaves. The only difference between them is the model topology. We found some branched networks may be organized into a complex topology such as a DAG \cite{murthy2016deep, huang2017multi, huang2017densely, mcgill2017deciding}. And, theoretically, can be designed as a tree-like structure. Hence, we deem that a branched network is a generalized expert NDT in terms of the model topology, and an expert NDT is actually a branched network with multi-exits. These approaches reduce the computation cost by adopting partial components of the network when performing inference on an input, and allow early exit when the network is already able to reach correct predictions. They are thus able to solve the wasteful overthinking of DNNs and lend evidence that decision branches for dynamic routing can be effective. However, the complex model topology requires prior expertise for the network design and may lead to more parameters during the training stage. With regard to interpretability, these decision branches, as specifically functionalized modules with explicit purposes, also increase "what" of the model, but other components remaining in the black box are still functionally ambiguous. 

In short, DT-inspired NNs lack the actual participation of DTs, so that they can not enjoy the same interpretability as NDTs. They can only offer limited “What” with weak decomposability. But from another perspective, these approaches generalize NDTs with regard to the employment of class hierarchies or the model topology. Besides, they may benefit from the absence of great constrains imposed by DTs and perform better in other aspects like performance and efficiency.

\subsection{Other Discussions and Considerations \label{sec:challenge&direction}}
This section concerns about the rest approaches that account for a few part of this survey. Despite the summaries and discussions on other aspects in previous taxonomic sections, we demonstrate that interpretability and other considerations are also partially touched in these techniques.

\subsubsection{\textbf{Saliency Methods}} They aim to identify which attributes of the input data are most relevant to a prediction or a latent representation \cite{fan2021interpretability}, belonging to post-hoc interpretability analysis. They partially learn the "Why" of the model, because the saliency indicates the critical parts of reasoning. Most saliency methods are tasked to obtain saliency maps, and an instance related to NDTs is \cite{li2019visualizing}. It traces the decision-making process of the DNDF \cite{kontschieder2015deep} and visualize saliency maps to understand which portion of the input influence it more. This survey touched saliency methods by introducing several axis-aligned DTs \cite{schmitz1999ann, schmitz1999combinatorial, wang2005integrated} in \autoref{sec:DT_Assist_NN}. They study the change of the prediction after perturbing or removing one attribute, which is limited to low-dimensional tasks or engineered features. We suggest that researchers resort to more complex saliency methods like gradient analysis \cite{simonyan2013deep, singla2019understanding, smilkov2017smoothgrad, sundararajan2017axiomatic}, class activation map (CAM) \cite{selvaraju2017grad, zhou2016learning} and apply them to design DTs or study NDTs.

\subsubsection{\textbf{Simulatability}} Simulatability is another criterion for interpretability. Unlike the "What" and "Why" which focus on decomposing the black box into transparent components, simulatability is dedicated to understanding the mechanism of the entire model at the top level in a unified theoretical framework \cite{fan2021interpretability, lipton2018mythos}. It is believed that a simpler model tends to have higher simulatability, so methods in \hyperref[sec:DT-interpreting-NN]{section 2.2} try to use simpler models to approximate the complex black-box model. If we consider enhancing simulatability in ante-hoc modeling, we are supposed to simplify the model. In \hyperref[sec:DT-regularization]{section 2.2.4} we introduced several approaches that use crafted regularization terms so as to simplify the decision boundaries of NNs. However, these methods are often adopted to serve post-hoc analysis, and more attempts for model facilitation need to be explored in the future.

\subsubsection{\textbf{Conditional Computation}} It is the central idea that indicates whether one model parallelizes or not. Typical DTs remain one of the most appealing algorithms in this field, where conditional computation is exploited to discard a gradually larger set of parameters and avoid performing the associated computation \cite{bengio2013deep}. But such a mutually exclusive condition can not be calculated in parallel and can hardly be applied to deep learning. Therefore, scholars resort to forming conditions in a distributed pattern. A common practice is to activate only some of the units in a NN in an input-dependent manner \cite{bengio2013deep, bengio2013estimating, bengio2015conditional}. They still use the indicator function in their "gating unit" that takes a hard binary decision for whether to turn off a unit in the hidden layer, but they allow this decision to be stochastic by estimating the gradient of a loss function with respect to the input of these non-smooth neurons. In view of NDTs, if we adopt fuzzy trees with soft inference during the training and testing phase, we can enjoy the parallel computation and distributed representation, but it does not inactivate any inconsequential neurons and performs full computation. On the contrary, the single-path hard inference carries out conditional computation with much less costs, but loses the capability of parallel computing because of the mutually exclusive condition. We suggest the future NDTs keep inferring by fuzzy NDTs, but should learn from the practices of \cite{bengio2013deep, bengio2013estimating, bengio2015conditional} such as using unbiased gradient estimation, in hope to realize stochastic inactivation of branches or other neural components so as to reduce the computation costs.

\section{Conclusion\label{sec:conclusion}}
This paper presents a comprehensive survey on the techniques of integrating NNs and DTs, namely neural trees. We first provide a throughout taxonomy and each kind of technique ends with a summary of its contributions and limitations. The sequence of introducing different techniques implies the gradual integration and the co-evolution of NNs and DTs, from separated cooperation or half integration to hybrid models that benefit from both. Afterward, we provide analysis for these techniques that focus on their interpertability. Finally, some extensional topics and promising future directions are discussed in hope to promote the future researches towards this field.

\bibliographystyle{ACM-Reference-Format}
\bibliography{sample-base}


\end{document}